\documentclass[10pt,twocolumn,letterpaper]{article}

\usepackage{iccv}              % To produce the CAMERA-READY version
\usepackage{amsmath,amsfonts,bm}

\def\eqref#1{equation~\ref{#1}}
\def\1{\bm{1}}

\def\vc{{\bm{c}}}

\def\vv{{\bm{v}}}

\DeclareMathAlphabet{\mathsfit}{\encodingdefault}{\sfdefault}{m}{sl}
\SetMathAlphabet{\mathsfit}{bold}{\encodingdefault}{\sfdefault}{bx}{n}

\DeclareMathOperator*{\argmin}{arg\,min}

\definecolor{iccvblue}{rgb}{0.21,0.49,0.74}
\usepackage[pagebackref,breaklinks,colorlinks,allcolors=iccvblue]{hyperref}

\usepackage{amsthm}
\usepackage{amsmath}
\usepackage{amssymb}
\usepackage{subcaption}
\usepackage{multirow}
\usepackage{algorithm}
\usepackage[switch]{lineno}

\usepackage{enumitem}
\setlist[itemize]{align=parleft,left=0pt..1em}

\usepackage{microtype}
\usepackage{graphicx}
\usepackage{booktabs}

\theoremstyle{definition}

\usepackage[noend]{algpseudocode}
\usepackage{pifont}

\def\confName{ICCV}
\def\confYear{2025}

\title{Heterogeneous Federated Learning with Prototype Alignment and Upscaling}

\author{Gyuejeong Lee\\
SAKAK Inc.\\
{\tt\small regulation.lee@sakak.co.kr}
\and
Jihwan Shin\\
SAKAK Inc.\\
{\tt\small jihwan@sakak.co.kr}
\and
Daeyoung Choi\\
The Cyber University of Korea\\
{\tt\small choidy@cuk.edu}
}

\begin{document}
\maketitle
\begin{abstract}
Heterogeneity in data distributions and model architectures remains a significant challenge in federated learning (FL). Various heterogeneous FL (HtFL) approaches have recently been proposed to address this challenge. Among them, prototype-based FL (PBFL) has emerged as a practical framework that only shares per-class mean activations from the penultimate layer. However, PBFL approaches often suffer from suboptimal prototype separation, limiting their discriminative power. We propose Prototype Normalization (ProtoNorm), a novel PBFL framework that addresses this limitation through two key components: Prototype Alignment (PA) and Prototype Upscaling (PU). The PA method draws inspiration from the Thomson problem in classical physics, optimizing global prototype configurations on a unit sphere to maximize angular separation; subsequently, the PU method increases prototype magnitudes to enhance separation in Euclidean space. Extensive evaluations on benchmark datasets show that our approach better separates prototypes and thus consistently outperforms existing HtFL approaches. Notably, since ProtoNorm inherits the communication efficiency of PBFL and the PA is performed server-side, it is particularly suitable for resource-constrained environments.
\end{abstract}    
\section{Introduction}
Federated learning (FL) enables collaborative model training across distributed clients without raw data sharing, demonstrating remarkable success in privacy-sensitive domains, including healthcare and finance applications \citep{zhang2021survey}. As a pioneering FL algorithm, FedAvg \citep{mcmahan2017communication} operates by aggregating local model parameters at a central server to create a single global model, which is then distributed back to clients for local training. However, its effectiveness is limited to scenarios with nearly uniform data distributions across homogeneous model architectures. Other traditional FL approaches based on model parameter sharing often face the same limitations. Data and model heterogeneity across clients is prevalent in practice, emerging as a fundamental challenge that demands novel solutions \citep{t2020personalized,li2022federated}.

To address these challenges, heterogeneous FL (HtFL) has emerged as a new paradigm that enables clients to employ different model architectures and handle non-uniform data distributions \citep{tan2022fedproto}. One line of approaches adapts parameter sharing by enabling clients to train architecture-specific subnets based on their available resources, addressing both system and statistical challenges \citep{diao2020heterofl,horvath2021fjord,zhu2022resilient,alam2022fedrolex}. However, these approaches inherently limit the degree of heterogeneity that can be accommodated, as they require architectural compatibility between shared components. Knowledge distillation (KD) based approaches have been proposed as an alternative that enables an entirely architecture-free federation to overcome this limitation. These approaches focus on knowledge sharing between the server and clients through various representations including logits \citep{jeong2018communication, li2019fedmd}, intermediate features \citep{tan2022fedproto, zhang2024fedtgp}, and auxiliary networks \citep{shen2020federated,zhu2021data}, offering greater flexibility in handling heterogeneous architectures.

Among the KD-based approaches, prototype-based FL (PBFL) has arisen as a particularly effective framework, operating by sharing prototypes—mean activations in the penultimate layer \citep{tan2022fedproto}. This approach offers several practical advantages: higher communication efficiency, enhanced privacy preservation, and implementation simplicity by sharing only feature vectors per class. Despite its advantages, PBFL suffers an inherent constraint in its optimization framework. The separation between global prototypes is solely determined by the weighted averaging of local prototypes, leading to suboptimal inter-prototype discrimination. While recent works like FedTGP \citep{zhang2024fedtgp} and FedFTP \citep{yin2025controlling} have addressed this through contrastive learning, the problem of optimal prototype separation remains an open challenge.

To address this prototype separability challenge, we propose a novel PBFL framework, Prototype Normalization (ProtoNorm). Our framework consists of two sequentially performed components: Prototype Alignment (PA) and Prototype Upscaling (PU). The PA method draws inspiration from the Thomson problem in classical physics, which seeks minimum energy configurations of $N$ electrons on a unit sphere. By implementing a novel gradient ascent-based Thomson problem solver, PA optimizes global prototype configurations on the unit sphere to maximize angular separation. The server incrementally adjusts each prototype's position to minimize hyperspherical energy until convergence. Following alignment, the PU method increases prototype magnitudes to enhance separation in Euclidean space—a critical step because training can converge to suboptimal local minima when global prototypes remain unit vectors after PA application.

Our motivational experiment on a synthetic spiral dataset visually demonstrates that standard PBFL approaches can fail to separate global prototypes adequately when dealing with non-independent and identically distributed (non-IID) data. In contrast, ProtoNorm achieves better separation of global prototypes even with non-IID data. We further demonstrate that prototype enlargement provides additional prototype separation benefits.
Extensive evaluations on popular multi-class benchmark datasets, including Tiny ImageNet, confirm that ProtoNorm consistently outperforms existing PBFL methods and other data-free HtFL approaches. Ablation studies further validate that PA and PU contribute significantly to performance improvements, with their combination delivering superior results compared to either component in isolation. We analyze the convergence of ProtoNorm by examining both the number of PA iterations required across FL rounds and the pairwise global prototype distances during these iterations, confirming stable convergence behavior.

Our contributions can be summarized as follows:
\begin{itemize}[itemsep=-0.25em, topsep=-0.25em]
\item We propose ProtoNorm, a novel PBFL framework that significantly improves inter-prototype discrimination through a two-phase approach: (1) PA: optimizing angular separation on the unit sphere, and (2) PU: upscaling prototypes to enhance separation in Euclidean space.
\item Through experiments on benchmark datasets, we demonstrate that ProtoNorm achieves optimal global prototype separation even with non-IID data, consistently outperforming existing HtFL approaches across various settings.
\item While maintaining the inherent communication efficiency of PBFL approaches, ProtoNorm performs server-side computation for PA and enhances privacy by eliminating the need for local class distribution information, making it well-suited for resource-constrained FL deployments.
\end{itemize}
\section{Related Work}
\subsection{Heterogeneous Federated Learning}
HtFL addresses challenges stemming from heterogeneity in the client model architecture. In this work, we focus on comparing our approach with KD-based HtFL. Existing KD-based HtFL methods can be classified according to the type of information they exchange.
Some approaches share model components or parameters. LG-FedAvg \citep{liang2020think} enables clients to share upper model layers while maintaining architecture-specific lower layers. FML \citep{shen2020federated} employs mutual distillation \citep{zhang2018deep} to train and share compact auxiliary models. FeDGen \citep{zhu2021data} exchanges a data generator to enhance generalization.
Other methods prioritize privacy by avoiding direct parameter sharing. FedDistill \citep{jeong2018communication} transmits globally averaged class-wise logits from clients to the server. However, this approach risks revealing the number of classes and logit distribution, creating potential privacy vulnerabilities. To address these concerns, FedProto \citep{tan2022fedproto} adopts a more privacy-preserving strategy by exchanging only local prototypes from the penultimate layer.
Recent works have further refined PBFL techniques to improve performance. FedTGP \citep{zhang2024fedtgp} and FedFTP \citep{yin2025controlling} enhance prototype separability through contrastive learning. Our work also builds upon PBFL and addresses prototype separability but takes a different approach by leveraging geometric insights inspired by the Thomson problem. Moreover, the computation overhead of the PA method scales only with the number of global prototypes, which equals the class count, and remains independent of the number of clients, thus making it particularly suitable for large-scale FL.

\subsection{Hyperspherical Approaches in Federated Learning}
In centralized machine learning, several studies have explored hyperspherical approaches, emphasizing that angular information is important for the semantics of deep networks, particularly in the computer vision domain \cite{chen2020angular,davidson2018hyperspherical,deng2019arcface,jing2021balanced,liu2021orthogonal,liu2021learning,liu2016large,liu2017deep,wang2018cosface,xu2019larger,liu2018learning}.
Among them, MHE, inspired by the Thomson problem, regularizes deep networks by minimizing hyperspherical energy to improve generalization. CoMHE \citep{lin2020regularizing} further extends MHE by leveraging projection mappings. HCR \citep{tan2022hyperspherical} applies regularization by projecting both classifier logits and feature embeddings onto their corresponding hyperspheres.
Hyperspherical approaches have also been actively applied to FL.
FedHP \cite{fonio2024fedhp} introduces a regularization term based on hyperspherical prototypical networks \cite{mettes2019hyperspherical} to uniformly partition the embedding space.
FedNH \citep{dai2023tackling} incorporates normalization layers to maintain hyperspherical uniformity and preserve semantic representations of class prototypes.
FedUV \citep{son2024feduv} employs hyperspherical uniformity to regularize representations, effectively simulating IID conditions with non-IID data.
Our ProtoNorm framework also applies a hyperspherical approach inspired by the Thomson problem but with key differences from previous work. While existing methods incorporate regularization terms jointly optimized during training, ProtoNorm optimizes global prototypes as a separate process on the server side. By leveraging the inherent advantages of PBFL, ProtoNorm supports heterogeneous models, unlike FedHP, FedNH, and FedUV. Moreover, this work verifies that angular information obtained through hyperspherical approaches can work synergistically with magnitude information.
\section{Problem Formulation and Motivation} \label{sec:problem}
This section outlines the problem formulation and the motivation behind our approach through an example.

\subsection{Problem Formulation} 
We examine an FL environment consisting of a central server connected to $M$ clients, each utilizing a distinct model architecture and maintaining its own private data distribution $P_i$ for classification tasks.
Following \cite{tan2022fedproto} and \cite{zhang2024fedtgp}, we partition each client's network $\bm{w}_i$ into two functional components: representation layers (feature extractor $f_i$ with parameters $\bm{\theta}_i$) that transform inputs from $\mathbb{R}^D$ into feature space $\mathbb{R}^d$, and a decision layer (classifier $g_i$ with parameters $\bm{\phi}_i$) that maps these features to outputs in $\mathbb{R}^K$, where $K$ denotes the number of classes. Therefore, for the sample $x$, we create a feature vector $\vc=f_i(\bm{\theta}_i;x)$.
The system objective is to minimize the average expected loss across all clients as follows:
\begin{equation}
\min_{\{\{\bm{\theta}_i, \bm{\phi}_i\}\}_{i=1}^{M}} \left\{ \frac{1}{M}\sum_{i=1}^{M}\mathbb{E}_{(x,y) \sim P_i}\left[\ell(\bm{\theta}_i, \bm{\phi}_i;x,y)\right] \right\}.
\end{equation}

In PBFL, clients exchange prototypes with the server instead of sharing model parameters. These prototypes are generated from the mean of decision layer activations for each class. For client $i$, the local prototype for class $j$ is defined as:
\begin{equation}
    \bar \vc_{i,j}^{L} = \frac{1}{n_{i,j}} \sum_{(x,y) \in \mathcal{D}_{i,j}} f_i(\bm{\theta}_i;x),
    \label{eq:local_prototype_aggregation}
\end{equation}
where $n_{i,j} = |\mathcal{D}_{i,j}|$ represents the number of samples from class $j$ on client $i$, and $\mathcal{D}_{i,j} \subseteq \mathcal{D}_i$ is the subset of client $i$'s local dataset containing only samples from class $j$. After receiving local prototypes from clients, the server creates global prototypes and sends them back to the clients. For each class $j$, the global prototype $\bar{\vc}_{j}^{G}$ is computed using weighted averaging as follows \citep{tan2022fedproto}:
\begin{equation}
\bar{\vc}_{j}^{G}=\frac{1}{\left|\mathcal{N}_j\right|} \sum_{i\in{\mathcal{N}_j}} \frac{n_{i,j}}{\sum_{i=1}^M n_{i,j}}\bar\vc_{i,j}^{L},
\label{eq:global_prototype_aggregation}
\end{equation}
where $\mathcal{N}_j$ is the set of clients with class $j$, and $\sum_{i=1}^M n_{i,j}$ is the total number of $j$-th class samples in the system. After receiving the global prototypes, each client trains its model using a combined loss function:
\begin{equation}
\Tilde{\mathcal{L}}_i(\bm{\theta}_i, \bm{\phi}_i) = \mathcal{L}_i(\bm{\theta}_i, \bm{\phi}_i) + \lambda\mathcal{R}_i,
\label{eq:loss_function}
\end{equation}
where $\mathcal{R}_i$ is the regularization term for prototype distance loss, and $\lambda$ is a hyperparameter controlling regularization strength. The term $\mathcal{R}_i$ is formulated as:
\begin{equation}
\mathcal{R}_i = \sum_{j} \rho(\bar{\vc}_{i,j}^{L}, \bar{\vc}_{j}^{G}),
\label{eq:regularization_term_fedproto}
\end{equation}
where the function $\rho(\cdot,\cdot)$ computes the Euclidean distance.
Unlike typical FL methods, PBFL approaches evaluate performance by calculating the distance between the decision layer activation and local prototype \citep{tan2022fedproto}:
\begin{equation}
\hat{y} = \argmin_j \| f_i(\bm{\theta}_i;x) - \bar{\vc}_{i,j}^{L} \|,
\label{eq:evaluation_metric}
\end{equation}
where $\|\cdot\|$ represents the Euclidean distance. 

\begin{figure}[b]
    \centering
    \begin{subfigure}{0.23\textwidth}
        \centering
        \includegraphics[width=\textwidth]{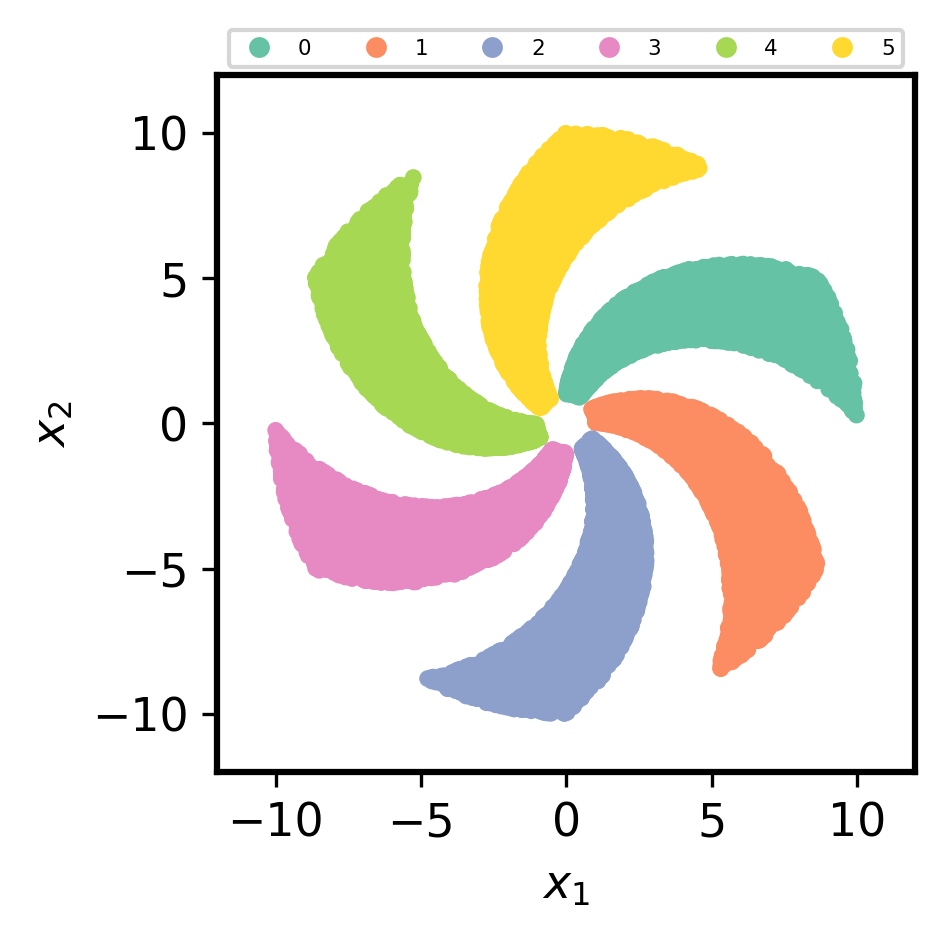}
        \caption{Spiral dataset}
        \label{fig:toy_dataset}
    \end{subfigure}
    \hfill
    \begin{subfigure}{0.234\textwidth}
        \centering
        \includegraphics[width=\textwidth]{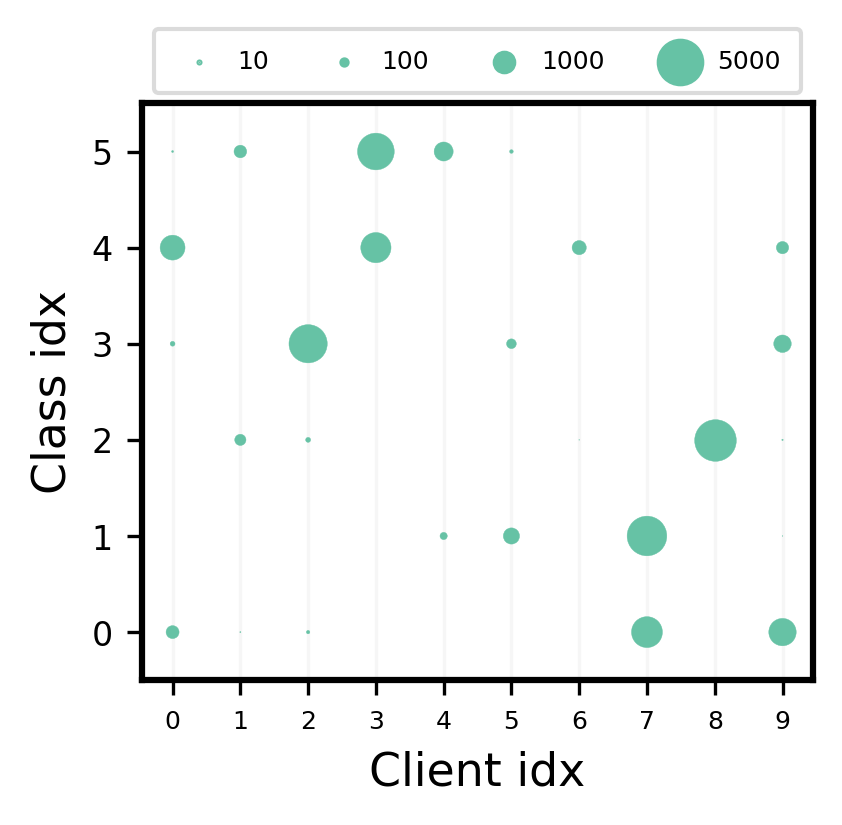} 
        \caption{Client data distribution}
        \label{fig:toy_data_distribution}
    \end{subfigure}
    \caption{Spiral dataset and non-IID data distribution across clients.}
\end{figure}

\begin{figure*}[t]
    \centering
    \begin{subfigure}{0.1645\textwidth}
        \centering
        \includegraphics[width=\textwidth]{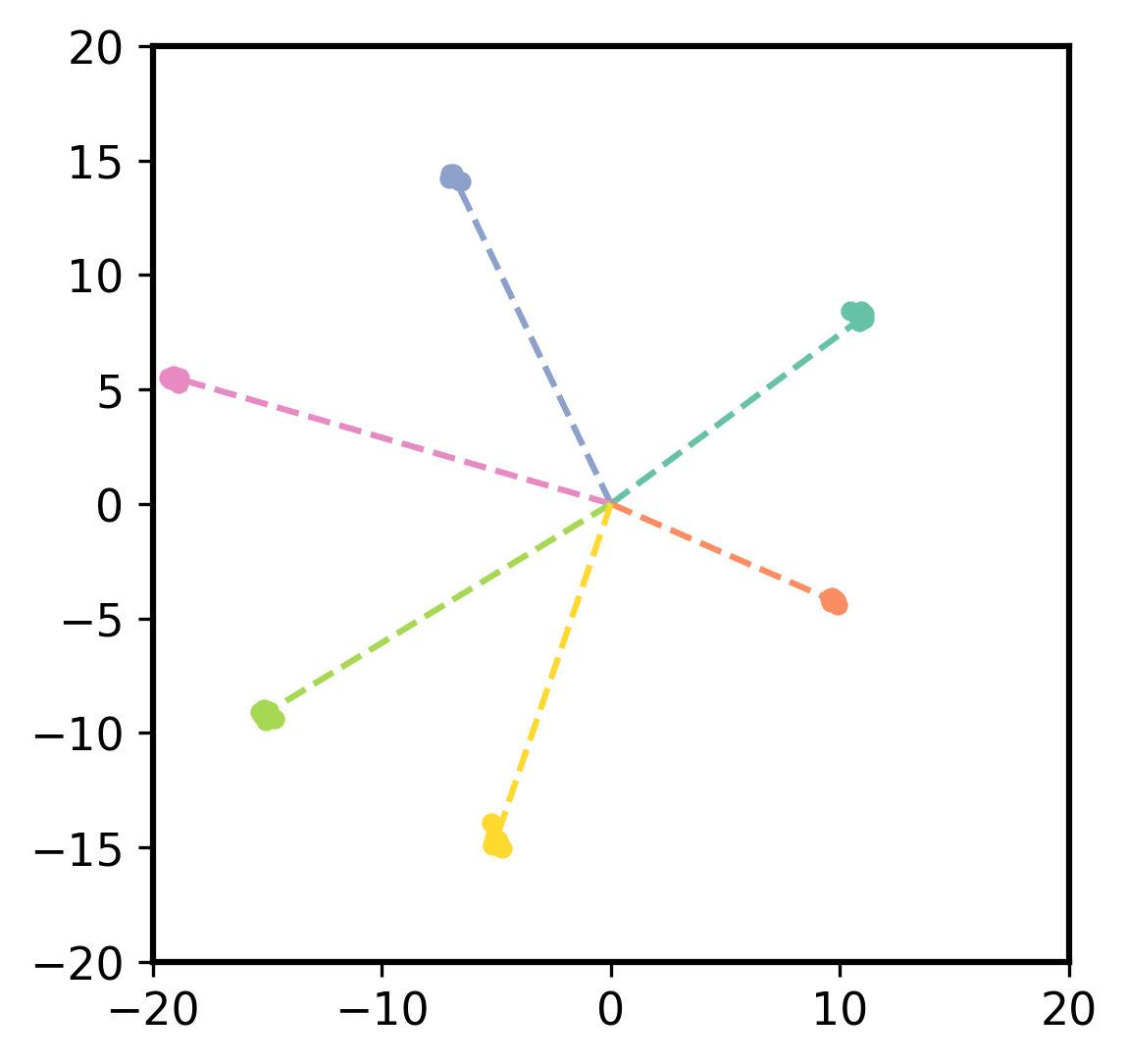}  
        \caption{FedProto \\(IID) }
        \label{fig:toy_fedproto_balanced}
    \end{subfigure}
    \hfill
    \begin{subfigure}{0.1555\textwidth}
        \centering
        \includegraphics[width=\textwidth]{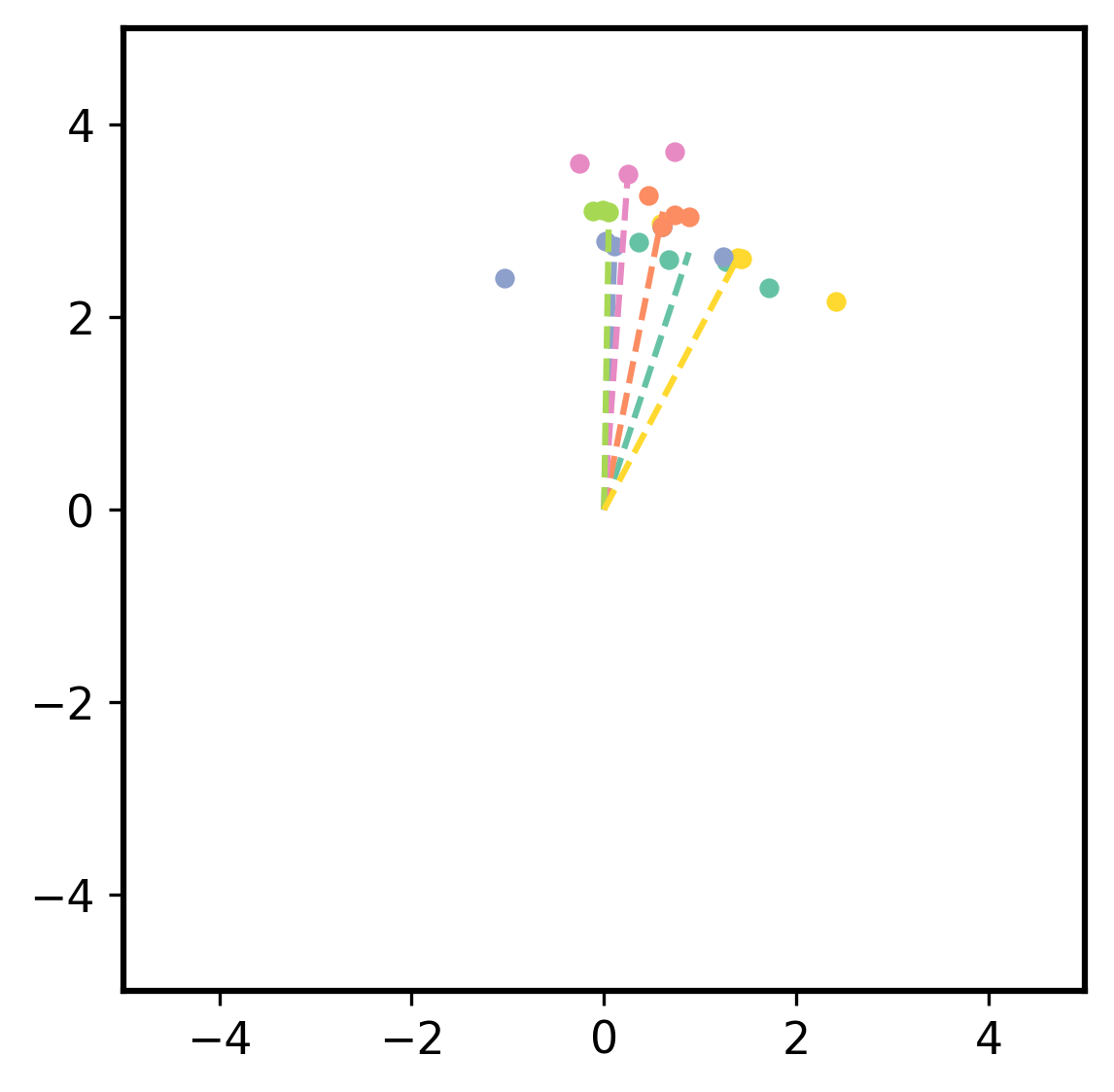}  
        \caption{FedProto \\(non-IID) }
        \label{fig:toy_fedproto_imbalanced}
    \end{subfigure}
    \hfill
    \begin{subfigure}{0.1615\textwidth}
        \centering
        \includegraphics[width=\textwidth]{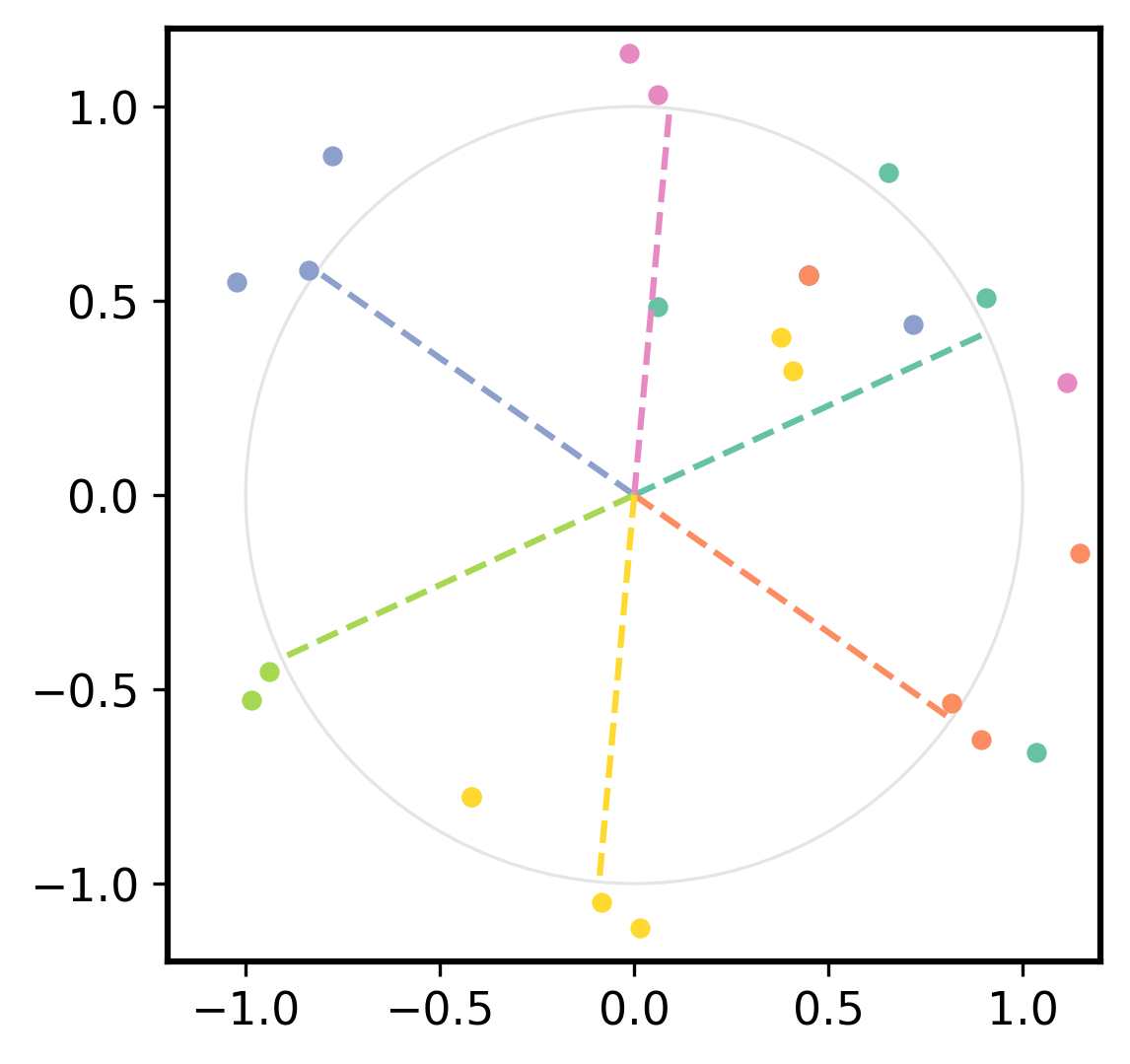}
        \caption{ProtoNorm \\(non-IID, $\gamma=1$)}
        \label{fig:toy_protonorm}
    \end{subfigure}
    \hfill
    \begin{subfigure}{0.16\textwidth}
        \centering
        \includegraphics[width=\textwidth]{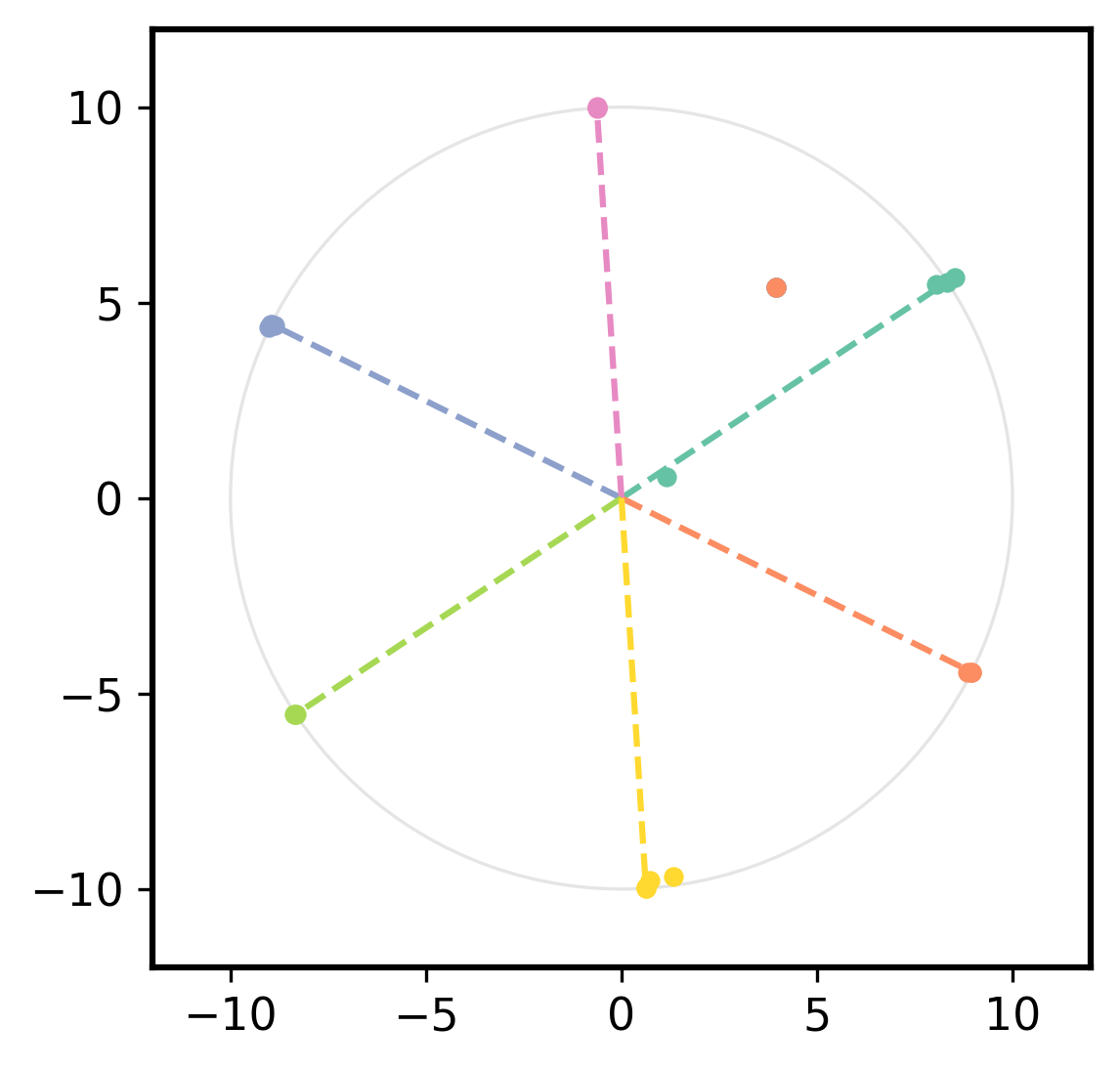}  
        \caption{ProtoNorm \\(non-IID, $\gamma=10$)}
        \label{fig:toy_protonorm_scaled}
    \end{subfigure}
    \hfill
    \begin{subfigure}{0.16\textwidth}
        \centering
        \includegraphics[width=\textwidth]{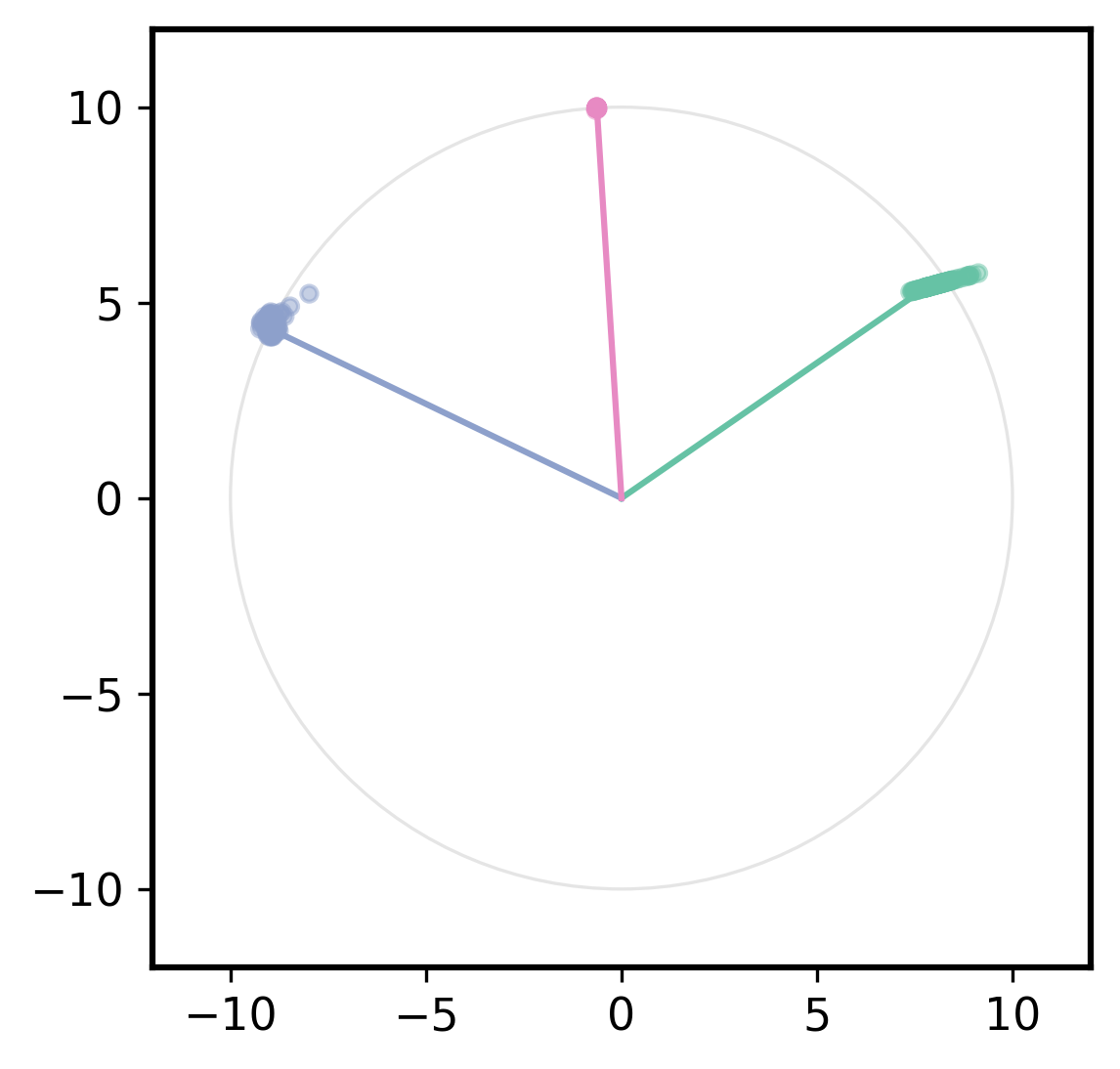}  
        \caption{Client 2 of  ProtoNorm (non-IID, $\gamma=10$)}
        \label{fig:toy_protonorm_scaled_clinet0}
    \end{subfigure}
    \hfill
    \begin{subfigure}{0.16\textwidth}
        \centering
        \includegraphics[width=\textwidth]{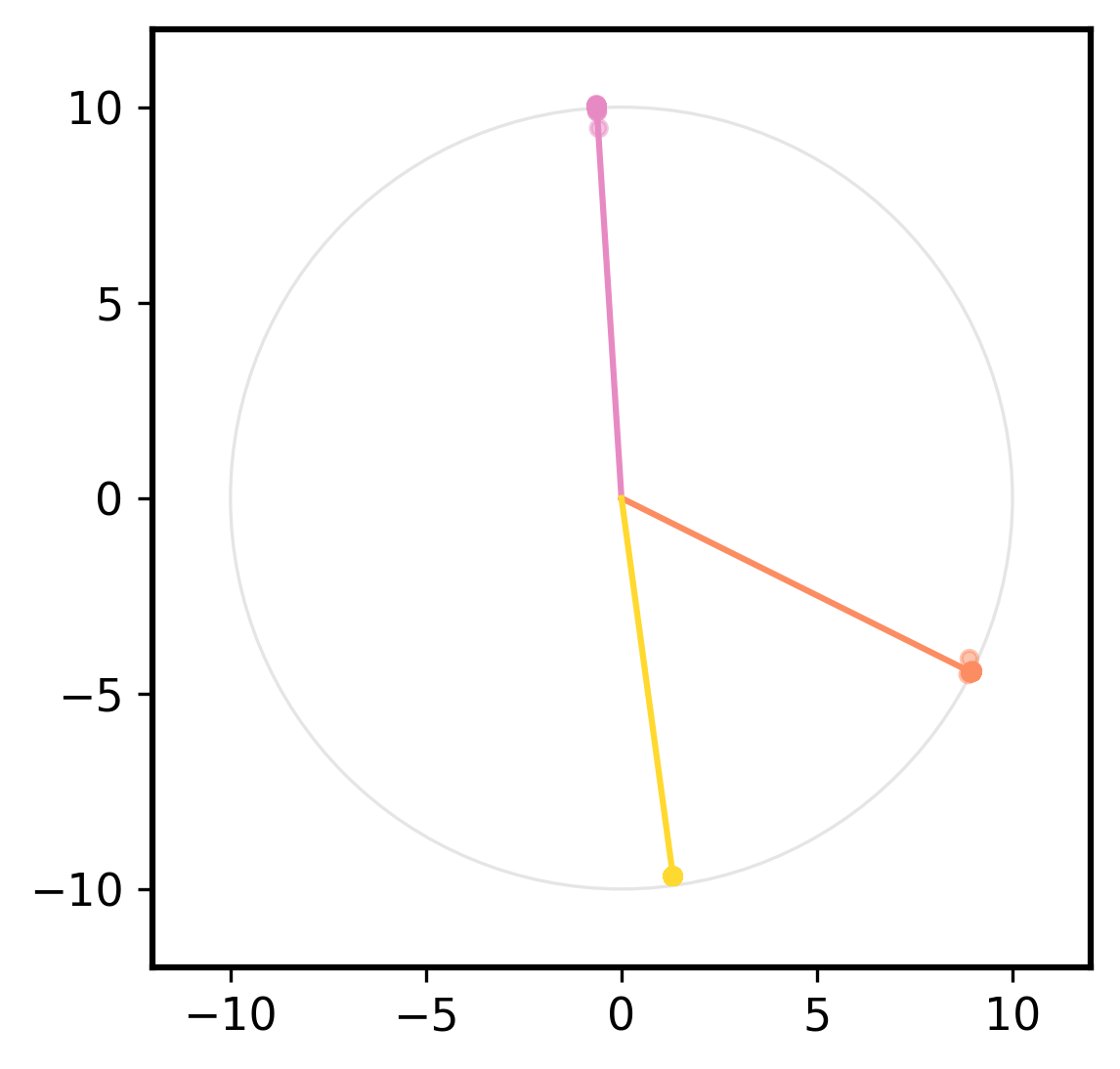}  
        \caption{Client 5 of  ProtoNorm (non-IID, $\gamma=10$)}
        \label{fig:toy_protonorm_scaled_clinet1}
    \end{subfigure}
    \caption{Visualization of prototypes and learned representations in two-dimensional feature space on the spiral dataset. Different colors denote different classes. (a)--(d) Global prototypes (dashed lines) and local prototypes (points) across different approaches. (e)--(f) Local prototypes (lines) and learned representations (points) of test samples of selected clients.} 
    \label{fig:toy_problem}
    \vspace{-15pt}
\end{figure*}

\subsection{Motivation} 
From the evaluation metric of PBFL in Eq. (\ref{eq:evaluation_metric}), it is evident that local prototypes should be maximally separated to guide class samples toward better separation effectively. This separation of local prototypes directly depends on the separation of global prototypes.
For optimal separation of vectors in Euclidean space, both angular distance and magnitude must be considered. However, FedProto does not explicitly optimize these two aspects. Instead, it simply aggregates local prototypes weighted by the number of class samples to generate global prototypes as described in Eq. (\ref{eq:global_prototype_aggregation}). The diverse data distributions and model architectures across clients result in substantially different local prototypes for the same classes, which can lead to suboptimal separation of global prototypes.

\noindent \textbf{A Motivational Example.} 
To demonstrate the necessity for separating global prototypes, we present an illustrative example using a synthetic spiral dataset in two dimensions (Figure~\ref{fig:toy_dataset}) \cite{dai2023tackling}. This dataset comprises six classes with 5000 points each, with its generation process detailed in Section A of the supplementary materials. The classifier employs a 5-layer multilayer perceptron with ReLU activation \citep{agarap2018deep}, where the dimensionality of the decision layer is set to two for visualization. Our experimental setup includes 10 clients under IID and non-IID data distribution scenarios. For the non-IID setting, data is distributed among clients following a Dirichlet distribution with parameter $\alpha=0.1$, effectively simulating practical heterogeneous FL scenarios (Figure~\ref{fig:toy_data_distribution}). This experimental design allows us to observe how prototype separation affects classification boundaries, providing clear visual evidence for our proposed approach.

We conducted experiments using FedProto and ProtoNorm for 200 FL rounds. Figure~\ref{fig:toy_fedproto_balanced} of FedProto shows that global prototypes (dashed lines) are nearly optimally separated, and local prototypes (points) are located near the global prototypes in the IID setting. However, under non-IID conditions, global prototypes lose this optimal separation as they align with clients possessing the majority of class samples as in Figure~\ref{fig:toy_fedproto_imbalanced}. Consequently, local prototypes become suboptimally separated as well.
In contrast, by applying the PA method of ProtoNorm shown in Figures~\ref{fig:toy_protonorm}, we distribute global prototypes uniformly on the unit sphere, resulting in better separated local prototypes even for non-IID settings. We further enhance prototype separation in Euclidean space by applying the PU method, which scales global prototypes by a factor of $\gamma=10$, as in Figure~\ref{fig:toy_protonorm_scaled}. With the better separation, learned representations of test samples (points) are positioned close to their respective local prototypes (lines), thereby ensuring higher classification performance as shown in Figure~\ref{fig:toy_protonorm_scaled_clinet0} and~\ref{fig:toy_protonorm_scaled_clinet1}.
These experiments reveal that effective prototype separation requires optimizing both angular distribution and magnitude components. Section B in the supplementary materials provides additional visualizations.
\section{Prototype Normalization} \label{sec:protonorm}
This section details the two phases of ProtoNorm.
\subsection{Phase 1: Prototype Alignment}
This phase is an extension of FedProto's prototype aggregation. The server first aggregates local prototypes from participating clients to generate global prototypes as FedProto. However, rather than using Eq. (\ref{eq:global_prototype_aggregation}), we employ a simple averaging approach to reduce privacy risks:
\begin{equation}
\bar{\boldsymbol{c}}_{j}^{G}=\frac{1}{\left|\mathcal{N}_j\right|} \sum_{i\in{\mathcal{N}_j}} \bar{\boldsymbol{c}}_{i,j}^{L}.
\label{eq:simple_aggregation}
\end{equation}
In the discussion section, we further discuss privacy enhancement using this approach. 
Subsequently, as an extension, the server aligns these prototypes to achieve maximal separation by applying the Prototype Alignment (PA) method. 
The PA is a gradient ascent-based optimization method that minimizes hyperspherical energy. It solves the Thomson problem of optimal point distribution on a hypersphere.

By analogy to the Thomson problem, minimum hyperspherical energy characterizes the equilibrium state of global prototype directional configurations \citep{liu2018learning}. We claim that prototypes arranged in a minimal energy configuration achieve better separation in $\mathbb{R}^d$. We omit the superscript $^G$ from the global prototype notation for notational convenience in this subsection.
Following \citep{liu2018learning}, we define the hyperspherical energy
functional for $K$ prototypes of $d$-dimension $\boldsymbol{\bar {C}}_K = \{\bar{\vc}_1, \cdots, \bar{\vc}_K \in \mathbb{R}^d\}$ as
\begin{equation}
\begin{aligned}
\boldsymbol{E}_{s,d}(\{\hat{\vc}_j\}_{j=1}^K) &= \sum_{j < k} h_s(\|\hat{\vc}_j - \hat{\vc}_k\|) \\
&= \begin{cases} 
\sum_{j < k} \|\hat{\vc}_j - \hat{\vc}_k\|^{-s}, & s>0 \\
\sum_{j < k} \log(\|\hat{\vc}_j - \hat{\vc}_k\|^{-1}), & s=0 
\end{cases},
\end{aligned}
\label{eq:original_optimization}
\end{equation}
where $h_s(\cdot)$ is a decreasing real-valued function and $\hat{\boldsymbol{c}}_j = \frac{\bar{\boldsymbol{c}}_j}{\|\bar{\boldsymbol{c}}_j\|}$
denotes the normalization of the $j$-th global prototype onto the unit hypersphere $\mathbb{S}^d = \{\boldsymbol{u} \in \mathbb{R}^{d}| \|\boldsymbol{u}\| = 1\}$. 
For notational efficiency, we define the set of normalized prototypes as $\hat{\boldsymbol{C}}_K = \{ \hat{\boldsymbol{c}}_1, \cdots, \hat{\boldsymbol{c}}_K \in \mathbb{S}^d\}$ and the corresponding hyperspherical energy as $\boldsymbol{E}_s = \boldsymbol{E}_{s,d}(\{\hat{\boldsymbol{c}}_j\}_{j=1}^K)$.
We employ Riesz $s$-kernels defined as $h_s(z) = z^{-s}$ with $s > 0$. 

The Thomson problem, which requires finding the minimum value of $\boldsymbol{E}_1$, has been proven to be NP-hard, making heuristic solutions necessary.
Mathematically, optimizing the original energy in Eq.~(\ref{eq:original_optimization}) is equivalent to minimizing its logarithmic form $\log \boldsymbol{E}_s$. To efficiently tackle this optimization challenge, we propose utilizing a surrogate energy function based on the lower bound of $\log \boldsymbol{E}_s$, which we derive through the application of Jensen's inequality:
\begin{equation}
\arg \min_{\hat{\boldsymbol{C}}_K} \{\boldsymbol{E}_{\log} := \sum_{j < k} \log(h_s(\|\hat{\vc}_j - \hat{\vc}_k\|))\}.
\end{equation}
Then, our surrogate energy function for $s=1$ using simplified notation is defined as follows:
\begin{equation}
\boldsymbol{E} = \sum_{j < k} \log  \frac{1}{{\left\| \hat{\boldsymbol{c}}_{j} - \hat{\boldsymbol{c}}_{k} \right\|}}. 
\label{eq:energy_term}
\end{equation}
During PA iterations, we first minimize this differentiable objective through gradient ascent with respect to the force $\boldsymbol{F}_j$ acting on $\hat{\boldsymbol{c}}_{j}$, defined as:
\begin{equation}
\boldsymbol{F}_j = -\nabla_{\hat{\boldsymbol{c}}_{j}} \boldsymbol{E} = \sum_{k=1, k\neq j}^{K} \frac{\hat{\boldsymbol{c}}_{j} - \hat{\boldsymbol{c}}_{k}}{\left\| \hat{\boldsymbol{c}}_{j} - \hat{\boldsymbol{c}}_{k} \right\|^2}.
\label{eq:force_term}
\end{equation}
This relationship follows the physical principle that force equals the negative gradient of potential energy. The negative sign in $-\nabla_{\hat{\boldsymbol{c}}_{j}} \boldsymbol{E}$ indicates that the force pushes prototypes away from configurations with higher energy toward those with lower energy, aligning with our goal of finding a minimal energy configuration. The server then updates the velocity vector $\boldsymbol{v}_j$, which determines the magnitude and direction of each prototype's movement:
\begin{equation}
\boldsymbol{v}_j^{(t)} = \mu \cdot \boldsymbol{v}_j^{(t-1)} + \eta^{(t)} \cdot \boldsymbol{F}_j^{(t)},
\label{eq:velocity}
\end{equation}
where $\mu$ is the momentum coefficient, and $\eta$ is the learning rate. 
Each prototype gradually adjusts its position using the updated velocity:
\begin{equation}
\tilde{\boldsymbol{c}}_j^{(t)} = \hat{\boldsymbol{c}}_j^{(t-1)} + \boldsymbol{v}_j^{(t)}.
\label{eq:prototype_update}
\end{equation}
After each update, we normalize the updated prototype:
\begin{equation}
\hat{\boldsymbol{c}}_j^{(t)} = \frac{\tilde{\boldsymbol{c}}_j^{(t)}}{\|\tilde{\boldsymbol{c}}_j^{(t)}\|}.
\label{eq:normalize_hat_c}
\end{equation}
The algorithm terminates after a predefined number of iterations or when meeting stopping criteria, ensuring optimization concludes when forces no longer induce significant changes in prototype positions as presented in Algorithm \ref{algo:pa}.

\begin{algorithm}[hb]
    \caption{Prototype Alignment}
    \hspace*{0.02in} {\bf Input:} $\bar{\vc}^{G}_{j}$, $j=1, \ldots, K$ \\
    \hspace*{0.02in} {\bf Output:} Optimized prototype vectors $\hat{\vc}^{G}_{j}$ \\
    \hspace*{0.02in} {\bf Server executes:}
    \begin{algorithmic}[1]
        \State Normalize $\bar{\vc}^{G}_{j}$ to create $\hat{\vc}^{G}_{j}$ by $\hat{\boldsymbol{c}}_j^{G} = \frac{\bar{\boldsymbol{c}}_j^{G}}{\|\bar{\boldsymbol{c}}_j^{G}\|}$
        \For{iteration $t = 1, \ldots, T$}
            \State Compute $\boldsymbol{F}_j$ using Eq. (\ref{eq:force_term}) 
            \State Update $\vv_j$ using {Eq. (\ref{eq:velocity})}
            \State Compute $\tilde{\vc}^{G}_{j}$ by updating $\hat{\vc}^{G}_{j}$ using Eq. (\ref{eq:prototype_update}) 
            \State Compute $\hat{\vc}^{G}_{j}$ by normalizing $\tilde{\vc}^{G}_{j}$ using Eq. (\ref{eq:normalize_hat_c})
            \If{stopping criterion met} 
                \State \textbf{break}
            \EndIf
        \EndFor
        \State \Return {$\hat{\vc}_{j}^{G}$}
    \end{algorithmic}
    \label{algo:pa}
\end{algorithm}

\subsection{Phase 2: Prototype Upscaling}
After the prototype alignment phase, we upscale the global prototypes on the client side.
Despite good angular separation, we observed that client performance deteriorates when global prototypes are normalized to unit vectors. 
This issue relates to local model optimization: as clients jointly optimize cross-entropy loss and the prototype distance loss, training can converge to suboptimal local minima when global prototype magnitudes are too small. 
A straightforward approach to address this issue is to upscale global prototypes as shown in Figure \ref{fig:toy_protonorm_scaled}--\ref{fig:toy_protonorm_scaled_clinet1}. By increasing their magnitudes, they can effectively guide local prototypes while preserving consistency with the activation scale of the feature extractor outputs. We implement Prototype Upscaling (PU) that can be achieved by applying a scaling hyperparameter $\gamma$ to global prototypes as follows:
\begin{equation}
   \mathcal{R}_i = \sum_{j} \rho(\bar{\boldsymbol{c}}_{i,j}^{L}, \gamma \cdot \hat{\boldsymbol{c}}_{j}^{G}).
   \label{eq:regularization_term}
\end{equation}
The comprehensive algorithm of ProtoNorm is provided in Algorithm \ref{algo:proto_norm}.

\begin{algorithm}[bh]
    \caption{Prototype Normalization}
    \label{alg:method}
    \hspace*{0.02in} {\bf Input:} $\mathcal{D}_i$, $\bm{w}_i$, $i=1, ..., M$ \\
    \hspace*{0.02in} {\bf Output:} Trained local models $\bm{w}_i$ for $i=1,...,M$ \\
    \hspace*{0.02in} {\bf Server executes:}
    \begin{algorithmic}[1]
        \State Initialize normalized prototype set $\left\{\hat{\vc}_{j}^{G}\right\}$ for all classes.
        \For{iteration $\tau = 1, \ldots, \mathcal{T}$}
            \State Sample a client subset $\mathcal{C}^{\tau}$
            \For{client $i \in \mathcal{C}^{\tau}$ in parallel}
                \State $\bar{\vc}_{i,j}^{L} \leftarrow$ LocalUpdate$\left(i, \hat{\vc}_{j}^{G}\right)$
            \EndFor
            \State Update $\bar{\vc}_{j}^{G}$ using Eq. (\ref{eq:simple_aggregation})  
            \State $\hat{\vc}_{j}^{G} \leftarrow$ Apply Algorithm \ref{algo:pa} with input $\bar{\vc}_{j}^{G}$

        \EndFor
    \end{algorithmic}
    \hspace*{0.02in} \\
    \hspace*{0.02in} {\bf LocalUpdate}$\left(i, \hat{\vc}_{j}^{G}\right)$:
    \begin{algorithmic}[1]
        \For{each local epoch}
            \For{batch $b \in \mathcal{B}_i$}
                \State Update model using the loss in Eq. (\ref{eq:loss_function}) and (\ref{eq:regularization_term})
            \EndFor
        \EndFor
        \State Compute $\bar{\vc}^{L}_{i,j}$ by Eq. (\ref{eq:local_prototype_aggregation})
        \State \Return {$\bar{\vc}^{L}_{i,j}$}
    \end{algorithmic}
    \label{algo:proto_norm}
\end{algorithm}
\begin{table*}[ht]
  \centering
  \setlength{\tabcolsep}{6.5pt}
  {\fontsize{9}{11}\selectfont
    \begin{tabular}{lrrrrrrrr}
      \toprule
      \multirow{2}{*}{Algorithm} & \multicolumn{3}{c}{Pathological setting}   & \multicolumn{4}{c}{Practical setting ($\alpha=0.1$)} \\
      \cmidrule(lr){2-4} \cmidrule(lr){5-8}
       & CIFAR-10 & CIFAR-100 & Tiny ImageNet & CIFAR-10 & CIFAR-100 & Flowers-102 & Tiny ImageNet\\
      \midrule
      LG-FedAvg         & 85.49 $\pm$ 0.05    & 53.51 $\pm$ 0.08   & 28.34 $\pm$ 0.08  & 86.97 $\pm$ 0.14  & 38.54 $\pm$ 0.21  & 43.88 $\pm$ 0.17   & 22.30 $\pm$ 0.37 \\
      FML               & 83.58 $\pm$ 0.06    & 51.76 $\pm$ 0.09   & 28.15 $\pm$ 0.09  & 86.59 $\pm$ 0.15  & 37.83 $\pm$ 0.03  & 40.39 $\pm$ 0.17   & 22.03 $\pm$ 0.12 \\
      FedKD             & 85.57 $\pm$ 0.05    & 53.12 $\pm$ 0.08   & 28.99 $\pm$ 0.07  & 87.10 $\pm$ 0.02  & 39.74 $\pm$ 0.42  & 42.06 $\pm$ 0.17   & 23.08 $\pm$ 0.17 \\
      FedDistill        & 85.53 $\pm$ 0.05    & 56.31 $\pm$ 0.07   & 30.40 $\pm$ 0.02  & 86.93 $\pm$ 0.12  & 39.52 $\pm$ 0.33  & 49.28 $\pm$ 0.12   & 22.98 $\pm$ 0.15 \\
      FedProto          & 80.55 $\pm$ 0.05    & 50.03 $\pm$ 0.09   & 25.20 $\pm$ 0.03  & 82.90 $\pm$ 0.46  & 29.97 $\pm$ 0.18  & 29.77 $\pm$ 0.15   & 13.30 $\pm$ 0.06 \\
      FedTGP            & 85.46 $\pm$ 0.05    & 50.57 $\pm$ 0.08   & 25.88 $\pm$ 0.04  & 86.32 $\pm$ 0.49  & 36.92 $\pm$ 0.16  & 43.95 $\pm$ 0.17   & 19.44 $\pm$ 0.12 \\
      \midrule
      ProtoNorm         & \textbf{88.40 $\pm$ 0.16}    & \textbf{64.18 $\pm$ 0.05}   & \textbf{38.75 $\pm$ 0.26}  & \textbf{88.56 $\pm$ 0.11}  & \textbf{47.41 $\pm$ 0.40}  & \textbf{53.83 $\pm$ 0.38}   & \textbf{31.20 $\pm$ 0.38} \\
      \bottomrule
    \end{tabular}
  }
  \caption{Classification accuracy (\%) on four benchmark datasets across varying data heterogeneity settings for heterogeneous model configurations.}
  \label{table:total_result}
  \vspace{-5pt}
\end{table*}

\begin{table*}[t]
    \centering
    \setlength{\tabcolsep}{7.pt}
    {\fontsize{9}{11}\selectfont
        \begin{tabular}{lrrrrrrrr}
            \toprule
            \multirow{2}{*}{Algorithm} & \multicolumn{3}{c}{Data heterogeneity ($\alpha$)} & \multicolumn{2}{c}{Decision layer dim. ($d$)} & \multicolumn{2}{c}{Number of clients ($M$)} \\
            \cmidrule(lr){2-4} \cmidrule(lr){5-6} \cmidrule(lr){7-8}
            & 0.01 & 0.5 & 1.0 & 64 & 1024 & 50 & 100 \\
            \midrule
            LG-FedAvg         & 66.75 $\pm$ 0.36  & 20.87 $\pm$ 0.18  & 15.84 $\pm$ 0.27  & 37.47 $\pm$ 0.17     & 39.50 $\pm$ 0.08 & 37.47 $\pm$ 0.19    & 35.28 $\pm$ 0.52 \\
            FML               & 64.29 $\pm$ 0.19  & 20.66 $\pm$ 0.20  & 15.58 $\pm$ 0.19  & 35.36 $\pm$ 0.19     & 39.06 $\pm$ 0.45 & 37.61 $\pm$ 0.12    & 35.69 $\pm$ 0.11 \\
            FedKD             & 66.08 $\pm$ 0.24  & 20.94 $\pm$ 0.21  & 16.01 $\pm$ 0.20  & 37.17 $\pm$ 0.40     & 40.06 $\pm$ 0.42 & 38.39 $\pm$ 0.15    & 35.85 $\pm$ 0.15 \\
            FedDistill        & 60.11 $\pm$ 0.21  & 22.28 $\pm$ 0.12  & 17.39 $\pm$ 0.39  & 37.08 $\pm$ 0.12     & 38.78 $\pm$ 1.11 & 40.70 $\pm$ 0.40    & 39.02 $\pm$ 0.04 \\
            FedProto          & 58.98 $\pm$ 0.31  & 12.88 $\pm$ 0.36  & 10.54 $\pm$ 0.24  & 25.86 $\pm$ 0.19     & 28.41 $\pm$ 0.11 & 29.36 $\pm$ 0.57    & 23.98 $\pm$ 0.34 \\
            FedTGP            & 64.02 $\pm$ 0.34  & 16.62 $\pm$ 0.28  & 11.97 $\pm$ 0.33  & 31.26 $\pm$ 0.69     & 37.24 $\pm$ 0.02 & 36.69 $\pm$ 0.15    & 35.22 $\pm$ 0.19 \\
            \midrule
            ProtoNorm           & \textbf{73.37 $\pm$ 0.23}  & \textbf{27.28 $\pm$ 0.07}  & \textbf{21.37 $\pm$ 0.08}  & \textbf{46.95 $\pm$ 0.44}     & \textbf{47.32 $\pm$ 0.15} & \textbf{43.98 $\pm$ 0.27}    & \textbf{41.06 $\pm$ 0.17} \\
            \bottomrule
        \end{tabular}
    }
    \caption{Classification accuracy (\%) on CIFAR-100 practical setting under varying data heterogeneity, decision layer dimensionality, and number of clients (client scalability).}
    \label{table:additional_result}
    \vspace{-5pt}
\end{table*}

\section{Experiments}  \label{sec:exp}
This section presents our experimental methodology, empirical results, and detailed analysis of convergence behavior.
\subsection{Experiment Setup}
\noindent {\bf Datasets and Model Architectures.}
We evaluate our approach using four datasets commonly employed in FL research: CIFAR-10/100 \citep{krizhevsky2009learning}, Flowers-102 \citep{nilsback2008automated}, and Tiny ImageNet \citep{le2015tiny}. For each dataset, we allocate 75\% for training and 25\% for testing. 
To replicate realistic non-IID data distributions among clients, we implement two distinct heterogeneity scenarios: (1) a pathological setting, where each client contains samples from only specific classes, and (2) a practical setting, where data is distributed according to a Dirichlet distribution with concentration parameter $\alpha = 0.1$ \citep{lin2020ensemble}.
We employ various lightweight models in our model heterogeneous setting, including ResNet-8 \citep{zhong2017deep}, EfficientNet \citep{tan2019efficientnet}, ShuffleNet v2 \citep{ma2018shufflenet}, and MobileNet v2 \citep{sandler2018mobilenetv2}. 
Each architecture incorporates a global average pooling layer, with the decision layer dimension fixed at $d = 512$.

\noindent {\bf Baselines and Evaluation Protocol.}
We evaluate the proposed methods against six state-of-the-art data-free HtFL algorithms: LG-FedAvg, FML, FedKD, FedDistill, FedProto, and FedTGP. For PBFL methods such as FedProto, FedTGP, and ProtoNorm, we evaluate model performance based on a similarity between local prototypes and decision layer activation as explained in Eq. (\ref{eq:evaluation_metric}). Following standard practice in the literature~\citep{t2020personalized, huang2021personalized, zhang2023fedala}, we adopt the same evaluation approach as FedTGP, measuring performance by the average test accuracy across the best-performing local models.

\noindent {\bf Federated Learning Setup.} Our FL setup comprises 20 clients, each participating in all 300 communication rounds. Client-side training uses a learning rate of 0.01, batch size of 32, and one local epoch per round. For the PA method, we set momentum coefficient $\mu=0.9$ and learning rate $\eta=0.1$ with 95\% decay every 10 iterations. We adopted $\lambda=1$ as in FedProto. Through grid search, we determined scale parameter $\gamma$ values of 100, 200, 200, and 400 for CIFAR-10, CIFAR-100, Flowers-102 and Tiny ImageNet, respectively. All results are reported as averages over three separate runs using distinct random seeds. Implementation details and code are available at: https://github.com/regulationLee/ProtoNorm

\subsection{Performance Results}
\noindent {\bf Overall Performance.}
Table \ref{table:total_result} demonstrates that ProtoNorm consistently outperforms the baselines across four datasets in model heterogeneous settings. Notably, ProtoNorm exhibits superior performance when dealing with many classes, where a clear separation of many global prototypes is required. 

\begin{table*}[th]
  \centering
  \setlength{\tabcolsep}{9pt}
  {\fontsize{9}{11}\selectfont
    \begin{tabular}{lrrrrrr}
      \toprule
      \multirow{2}{*}{Algorithm} & \multicolumn{3}{c}{4-layer CNN} & \multicolumn{3}{c}{ResNet-18} \\
      \cmidrule(lr){2-4} \cmidrule(lr){5-7}
       & CIFAR-100 & Flowers-102 & Tiny ImageNet& CIFAR-100 & Flowers-102 & Tiny ImageNet \\
      \midrule
      FedNH & 32.68 $\pm$ 0.61 & 36.61 $\pm$ 1.29 & 16.50 $\pm$ 0.07 & 37.72 $\pm$ 1.23 & 47.61 $\pm$ 1.09 & 22.62 $\pm$ 0.70 \\
      FedUV & 46.10 $\pm$ 0.21 & 50.48 $\pm$ 0.70 & 27.38 $\pm$ 0.07 & 48.33 $\pm$ 0.19 & 55.36 $\pm$ 0.32 & 30.18 $\pm$ 0.06 \\
      \midrule
      ProtoNorm & \textbf{48.25 $\pm$ 0.12} & \textbf{51.21 $\pm$ 0.54} & \textbf{29.35 $\pm$ 0.05} & \textbf{50.57 $\pm$ 0.21} & \textbf{60.28 $\pm$ 0.49} & \textbf{31.69 $\pm$ 0.32} \\
      \bottomrule 
    \end{tabular}
  }
  \caption{Classification accuracy (\%) on three benchmark datasets across two model architectures for homogeneous model settings.}
  \label{table:homo_result}
\end{table*}

\begin{table}[t]
    \centering
    \setlength{\tabcolsep}{4pt}
    {\fontsize{9}{11}\selectfont
        \begin{tabular}{lrrrrrr}
            \toprule
            \multirow{2}{*}{Dataset} & \multicolumn{6}{c}{Scaling factor ($\gamma$)} \\ 
            \cmidrule(lr){2-7}
            & 1 & 200 & 400 & 600 & 800 & 1000\\
            \midrule             
            CIFAR-100 & 29.71 & \textbf{47.41} & 46.89 & 46.63 & 46.66 & 45.72 \\
            Flowers-102 & 30.24 & \textbf{53.83} & 53.37 & 50.79 & 50.11 & 49.43 \\
            Tiny ImageNet & 15.27 & 30.59 & \textbf{31.20} & 30.61 & 29.66 & - \\
            \bottomrule
        \end{tabular}
    }
    \caption{Classification accuracy (\%) of ProtoNorm on CIFAR-100 across varying $\gamma$. `-' indicates the model failed to converge.}
    \label{table:protonorm_scaling}
\end{table}

\noindent {\bf Data Heterogeneity.}
We also conduct experiments to evaluate ProtoNorm's sensitivity to data heterogeneity. We vary data distributions using the $\alpha$ parameter of the Dirichlet distribution. This heterogeneity decreases as $\alpha$ increases, leading to a more balanced data distribution across classes for each client. Table~\ref{table:additional_result} shows that ProtoNorm consistently outperforms the baselines regardless of the data heterogeneity level, demonstrating our approach's robustness.

\noindent {\bf Decision Layer Dimensionality.}
The performance of deep networks is generally affected by decision layer dimensionality. Since PBFL methods share class prototypes derived from the decision layer, their performance can be particularly sensitive to this parameter. We evaluate this dimensional sensitivity across methods, with results presented in Table \ref{table:additional_result}. ProtoNorm consistently outperforms competing approaches across all tested dimensions. Notably, while the performance of other methods deteriorates with decreasing dimensionality, ProtoNorm maintains superior results even in low dimensionality.

\noindent {\bf Scalability.}
Table~\ref{table:additional_result} shows that the performance of all methods degrades as the number of clients increases, an expected challenge in FL. However, ProtoNorm consistently outperforms the baselines across different client scales, demonstrating superior robustness to this scaling challenge. 

\noindent {\bf Homogeneous Model Setting.}
Although primarily designed for model heterogeneous settings, ProtoNorm was compared with two model homogeneous methods that utilize hyperspherical uniformity. As shown in Table~\ref{table:homo_result}, ProtoNorm consistently outperforms FedNH and FedUV across all tested architectures (4-layer CNN and ResNet-18) and datasets (CIFAR-100, Flowers-102, and Tiny ImageNet). These results demonstrate the effectiveness of our approach in creating discriminative prototypes. Further advantages of ProtoNorm over these methods are discussed in Section~\ref{sec:discussion}.

\subsection{Ablation Study}
\noindent {\bf Effect of Prototype Alignment.} 
\noindent 
The effectiveness of the PA method is verified by comparing the performance of FedProto and ProtoNorm. As consistently shown in Tables \ref{table:total_result} and \ref{table:additional_result}, ProtoNorm outperforms FedProto when applying the PU. However, when the PU is not applied, there is no noticeable performance gain, as evidenced by comparing the accuracy of ProtoNorm with $\gamma=1$ (Table \ref{table:protonorm_scaling}) against that of FedProto with $\gamma=1$ (Table \ref{table:pbfl_scaling}) for CIFAR-100. This result suggests that prototype alignment works significantly better with prototype upscaling.

\begin{table}[t]
    \centering
    \setlength{\tabcolsep}{4pt}
    {\fontsize{9}{11}\selectfont
        \begin{tabular}{llrrrrrrr}
            \toprule
            \multirow{2}{*}{Dataset} & \multirow{2}{*}{Algorithm} & \multicolumn{5}{c}{Scaling factor ($\gamma$)} \\
            \cmidrule(lr){3-7}
             &  & 1.0 & 1.1 & 1.2 & 1.3 & 1.4 \\
            \midrule
            \multirow{2}{*}{CIFAR-10} & FedProto & 82.90 & \textbf{83.23} & 63.43 & 45.02 & -  \\
            & FedTGP & 86.32 & 87.11 & \textbf{87.16} & 86.83 & -  \\
            \midrule
            \multirow{2}{*}{CIFAR-100} & FedProto & 29.97 & \textbf{30.52} & 18.01 & - & -  \\
            & FedTGP & 36.92 & 44.55 & \textbf{47.33} & 45.10 & 44.84 \\
            \bottomrule
        \end{tabular}
    }
    \caption{Classification accuracy (\%) of other PBFL methods across varying $\gamma$. `-' indicates the model failed to converge.}
    \label{table:pbfl_scaling}
\end{table}

\noindent {\bf Effect of Prototype Upscaling.} 
\noindent 
To investigate the synergistic relationship between the PA and PU methods, we vary the scaling factor $\gamma$ on CIFAR-100, Flowers-102, and Tiny ImageNet as shown in Table \ref{table:protonorm_scaling}.  
The results clearly demonstrate that the scaling factor significantly influences performance. For instance, comparing the performance at the optimal $\gamma$ (highlighted in bold) with $\gamma=1$ reveals substantial improvements, suggesting that prototype upscaling works synergistically with prototype alignment.

From the study on ProtoNorm's scaling factor (Table \ref{table:protonorm_scaling}), one might question whether scaling could similarly improve the performance of other PBFL methods such as FedProto and FedTGP. To investigate this hypothesis, we conducted experiments with varying scaling factors on these methods using CIFAR-10 and CIFAR-100 datasets, as shown in Table~\ref{table:pbfl_scaling}.
Indeed, we found that their performance can be significantly improved by adjusting the scaling factor. For instance, FedTGP's accuracy on CIFAR-100 shows a remarkable increase from 36.92\% to 47.33\% with appropriate scaling.
However, these methods exhibit high sensitivity to the scaling factor, making them challenging to optimize effectively. Moreover, even with optimal scaling, their performance consistently remains below that of ProtoNorm.

\begin{figure}[tbh]
    \centering
    \begin{subfigure}{0.223\textwidth}
        \centering
        \includegraphics[width=\textwidth]{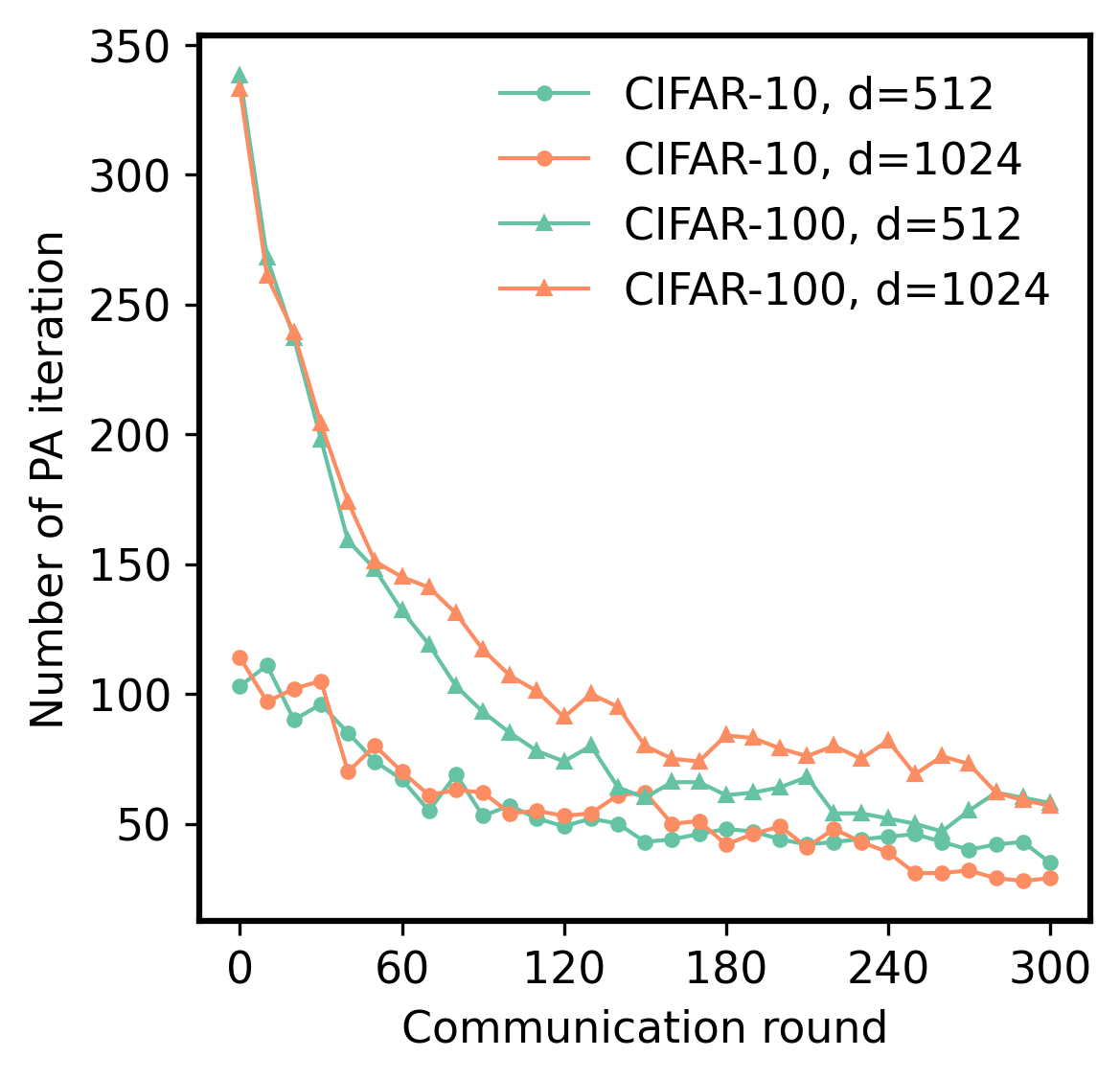}
        \caption{Number of PA iteration}
        \label{fig:convergence_pa_iteration}
    \end{subfigure}
    \begin{subfigure}{0.22\textwidth}
        \centering
        \includegraphics[width=\textwidth]{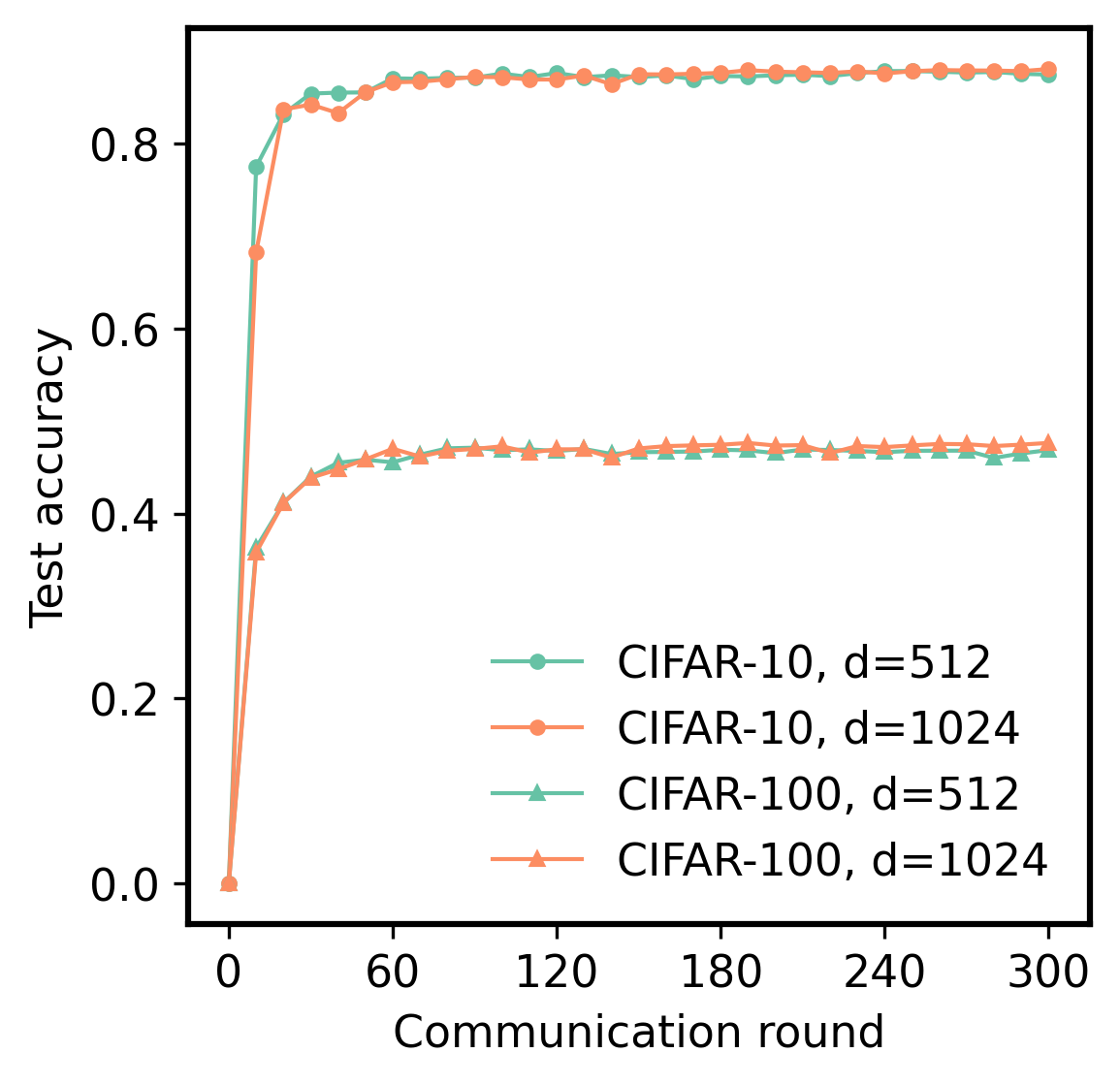} 
        \caption{Test accuracy}
        \label{fig:convergence_accuracy}
    \end{subfigure}
    % \vspace{-5pt}
    \caption{Convergence analysis.} 
    \label{fig:convergence}
\end{figure}

\begin{figure}[tbh]
    \centering
    \begin{subfigure}{0.2255\textwidth}
        \centering
        \includegraphics[width=\textwidth]{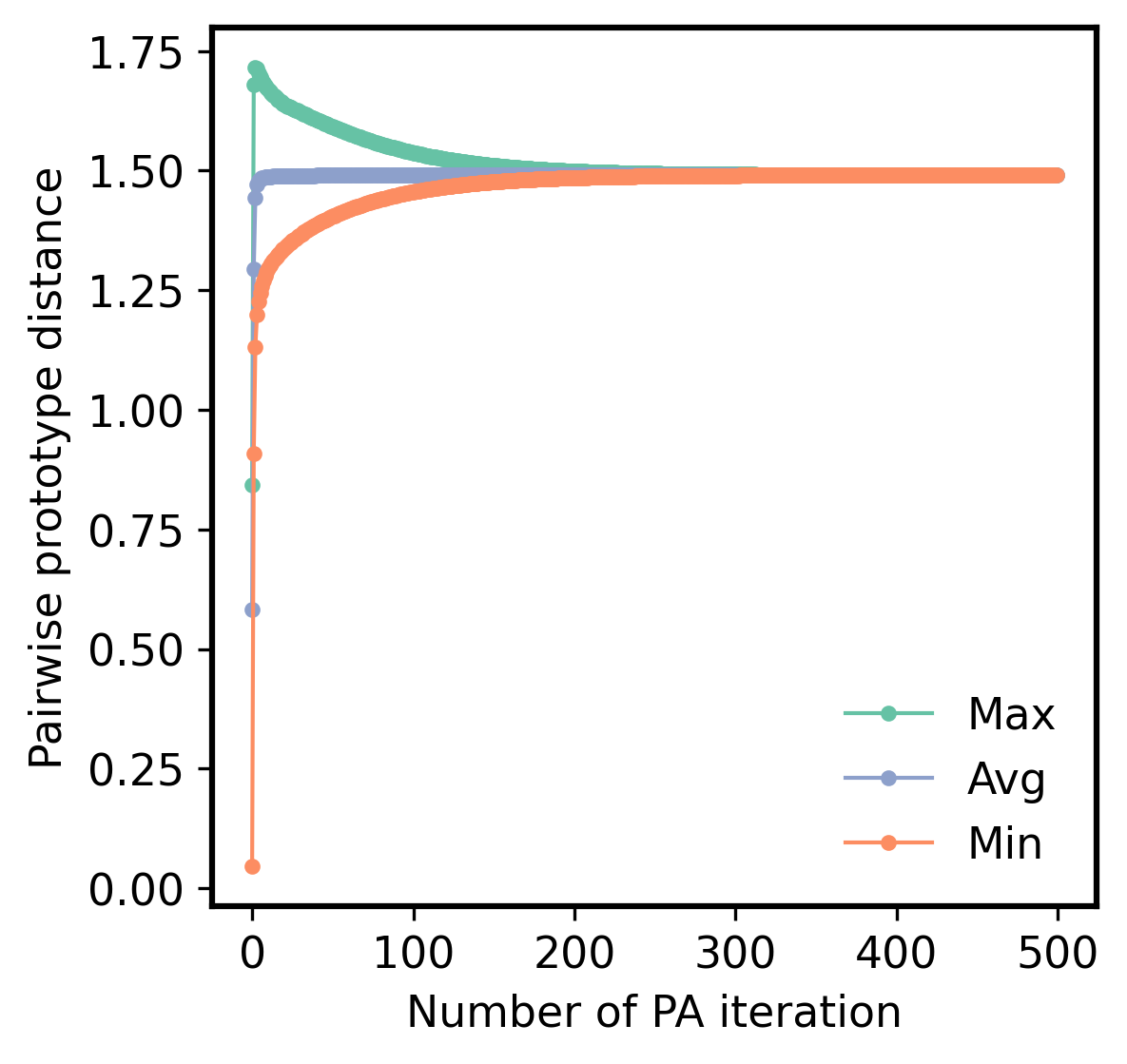}
        \caption{CIFAR-10}
        \label{fig:distance_cifar10_512}
    \end{subfigure}
    \begin{subfigure}{0.22\textwidth}
        \centering
        \includegraphics[width=\textwidth]{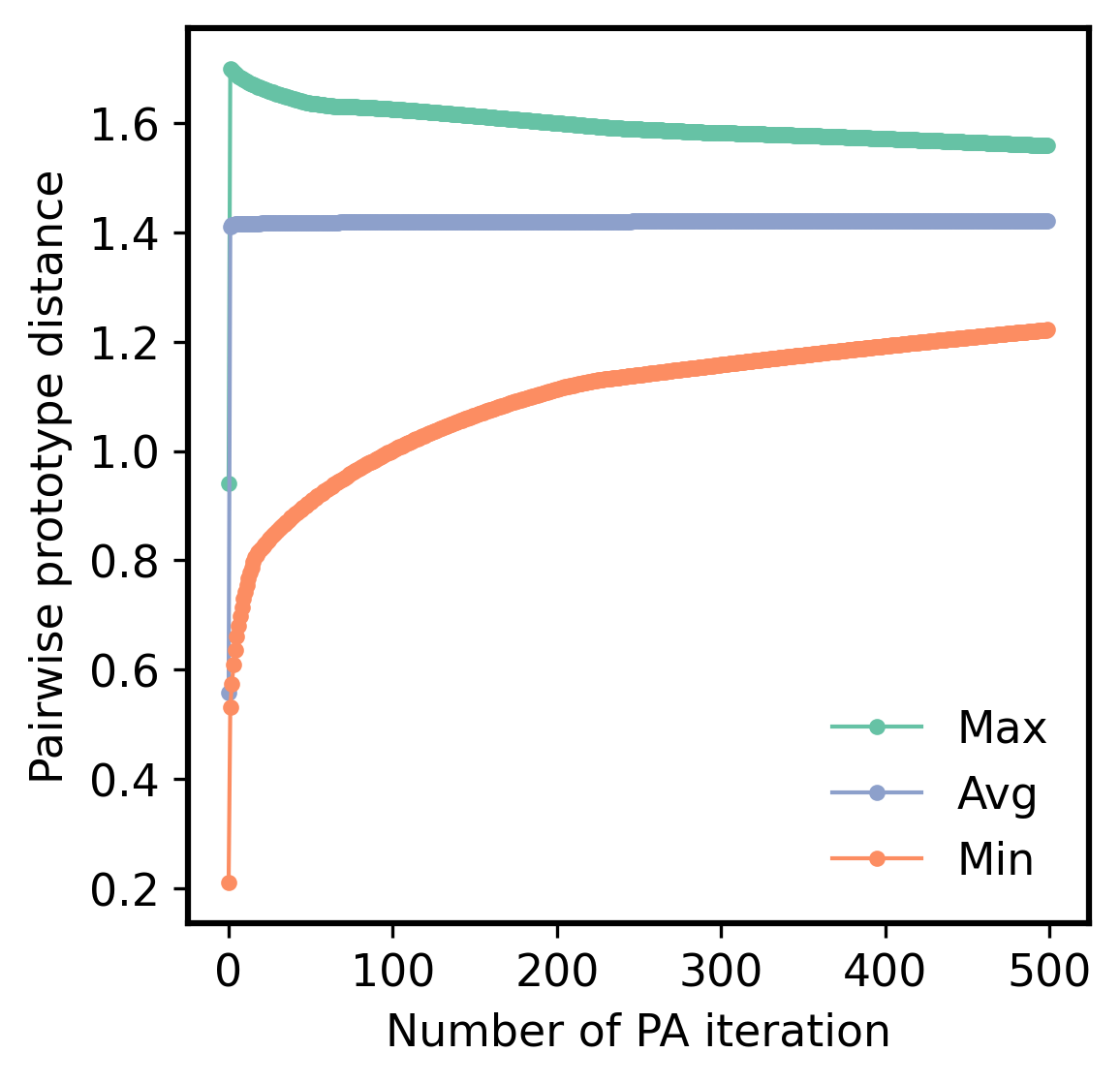} 
        \caption{CIFAR-100}
        \label{fig:distance_cifar100_512}
    \end{subfigure}
    % \vspace{-5pt}
    \caption{Pairwise global prototype distance during PA iterations.} 
    \label{fig:distance_512}
    \vspace{-10pt}
\end{figure}

\subsection{Convergence Analysis}
Our approach preserves both prototype aggregation and regularization in the local training of FedProto, maintaining the same non-convex convergence rate as FedProto. The PA method's convergence per communication round is guaranteed through momentum and learning rate decay. Therefore, this study focuses on analyzing the empirical convergence behaviors of PA and local training during whole FL rounds across various experimental settings.

Figure \ref{fig:convergence_pa_iteration} illustrates the decreasing number of PA iterations across 300 communication rounds. The PA process terminates when 10 consecutive changes of $\boldsymbol{F}_{j}$ fall below the threshold $\epsilon$. As expected, we observe a gradual reduction in required PA iterations throughout the FL process. Notably, CIFAR-100 exhibits a slower decrease rate compared to CIFAR-10, which is consistent with its more complex class structure containing 10$\times$ more categories. This convergence in PA iterations correlates with test accuracy, which converges smoothly regardless of dataset and prototype dimensionality, as shown in Figure \ref{fig:convergence_accuracy}.

Additionally, Figure \ref{fig:distance_512} provides geometric insight into the convergence properties of our approach. The line plots illustrate how pairwise global prototype distances evolve during a single PA phase at the first FL round. For both CIFAR-10 and CIFAR-100, the average distance between global prototypes approaches the expected distance of $\sqrt{2}$ in the 512-dimensional prototype space.
Especially when the class count $d$ is small relative to the dimensionality (Figure \ref{fig:distance_cifar10_512}), we observe that prototypes achieve near-optimal distribution. This occurs because prototypes can position themselves nearly orthogonal to each other. In contrast, when the class count $d$ is large, as in CIFAR-100, prototypes do not reach optimal distribution. Regardless, our experimental results show significant performance improvements for CIFAR-100 and even for datasets having more than 100 classes.
\begin{figure}[tbh]
    \centering
    \includegraphics[width=0.42\textwidth]{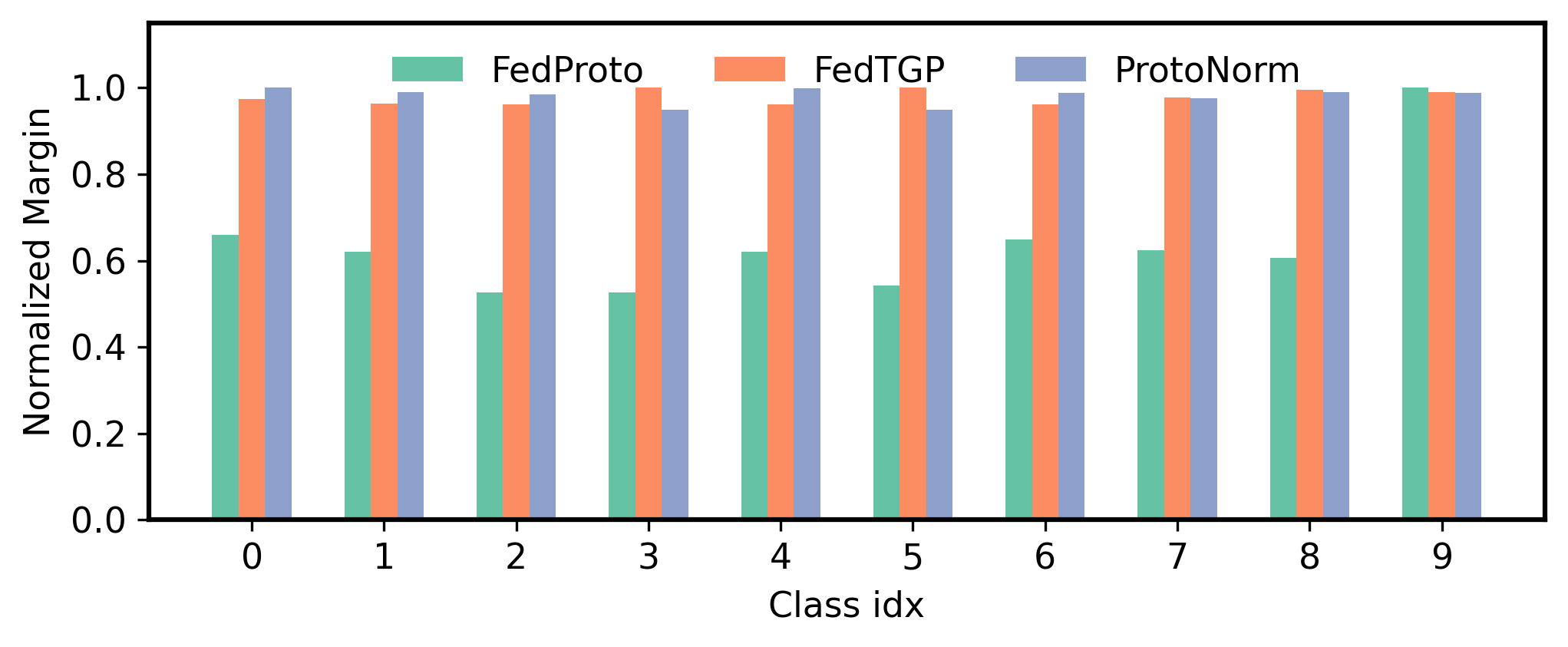}
    \vspace{-5pt}
    \caption{Comparison of normalized global prototype margins.}
    \label{fig:margin}
    \vspace{-5pt}
\end{figure}

\section{Discussion} \label{sec:discussion}
\subsection{Prototype Separability}
Following FedTGP \cite{zhang2024fedtgp}, we define the global prototype margin as the minimal Euclidean distance between the prototype of a specific class and prototypes of all other classes. The maximum margin represents the largest global prototype margin across all classes. We normalize all margin values to the maximum margin for each method in Figure \ref{fig:margin} to remove magnitude effects. In FedTGP and ProtoNorm, the normalized global prototype margin demonstrates similar values, indicating evenly distributed inter-prototype distances, which aligns with our observations in Figure \ref{fig:distance_512}. In contrast, FedProto exhibits irregular prototype distributions, suggesting suboptimal separation in the embedding space.

\subsection{Communication and Computational Efficiency}
Like other PBFL approaches, ProtoNorm maintains communication efficiency advantages over model homogeneous methods (FedNH and FedUV) and other HtFL methods by transmitting only class prototypes. A detailed analysis of this efficiency gain is provided in the supplementary materials. Notably, ProtoNorm offers a practical advantage over FedTGP in terms of client scalability. The computational overhead of FedTGP scales with client count, as local prototypes are required to train the model for server-side contrastive learning. In contrast, ProtoNorm avoids this computational burden by only using averaged global prototypes. Although one might be concerned about the computational cost of pairwise operations in the PA method, our implementation leverages modern tensor operations that enable efficient execution. These advantages make ProtoNorm particularly well-suited for large-scale FL deployments in resource-constrained environments.

\subsection{Privacy Enhancement}
The weighted averaging in Eq. (\ref{eq:global_prototype_aggregation}) necessitates that clients transmit their class distribution information $n_{i,j}$ to the server. This transmission presents a privacy vulnerability, as these distribution patterns may reveal sensitive characteristics of the client datasets \cite{yi2023fedgh,zhang2024fedtgp}. 
In contrast, the PA method circumvents this concern by achieving prototype separation on the unit sphere through angular relationships alone, eliminating any requirement for class distribution information.

{
    \small
    \bibliographystyle{ieeenat_fullname}
    \bibliography{main}
}

\clearpage
\setcounter{page}{1}
\maketitlesupplementary

\section*{A. Spiral Dataset Generation}
\label{sec:illustration_comparison}
The spiral dataset consists of six distinct spiral-shaped clusters, each representing one class. 
For each class $k \in \{0, \cdots, 5\}$, we generate 5000 data points as $\mathcal{C}_k = \{(x_{k,i}, y_{k,i}) \mid x_{k,i} = r_i \sin \omega_{k,i}, y_{k,i} = r_i \cos \omega_{k,i}, i \in [5000]\}$, where:
$$r_i = 1 + (i - 1)\frac{9}{4999} \text{ and } \omega_{k,i} = \frac{k}{3}\pi + (i - 1)\frac{k}{3 \times 4999}\pi + b_i$$ 
for all $i \in [5000]$. The term $b_i \sim \mathcal{N}_1(0, 1)$ introduces Gaussian noise to create realistic decision boundaries between classes.

\section*{B. Visualizations for the Spiral Dataset}  
\begin{figure}[htb]
    \centering
    \begin{subfigure}[b]{0.115\textwidth}
        \includegraphics[width=\textwidth]{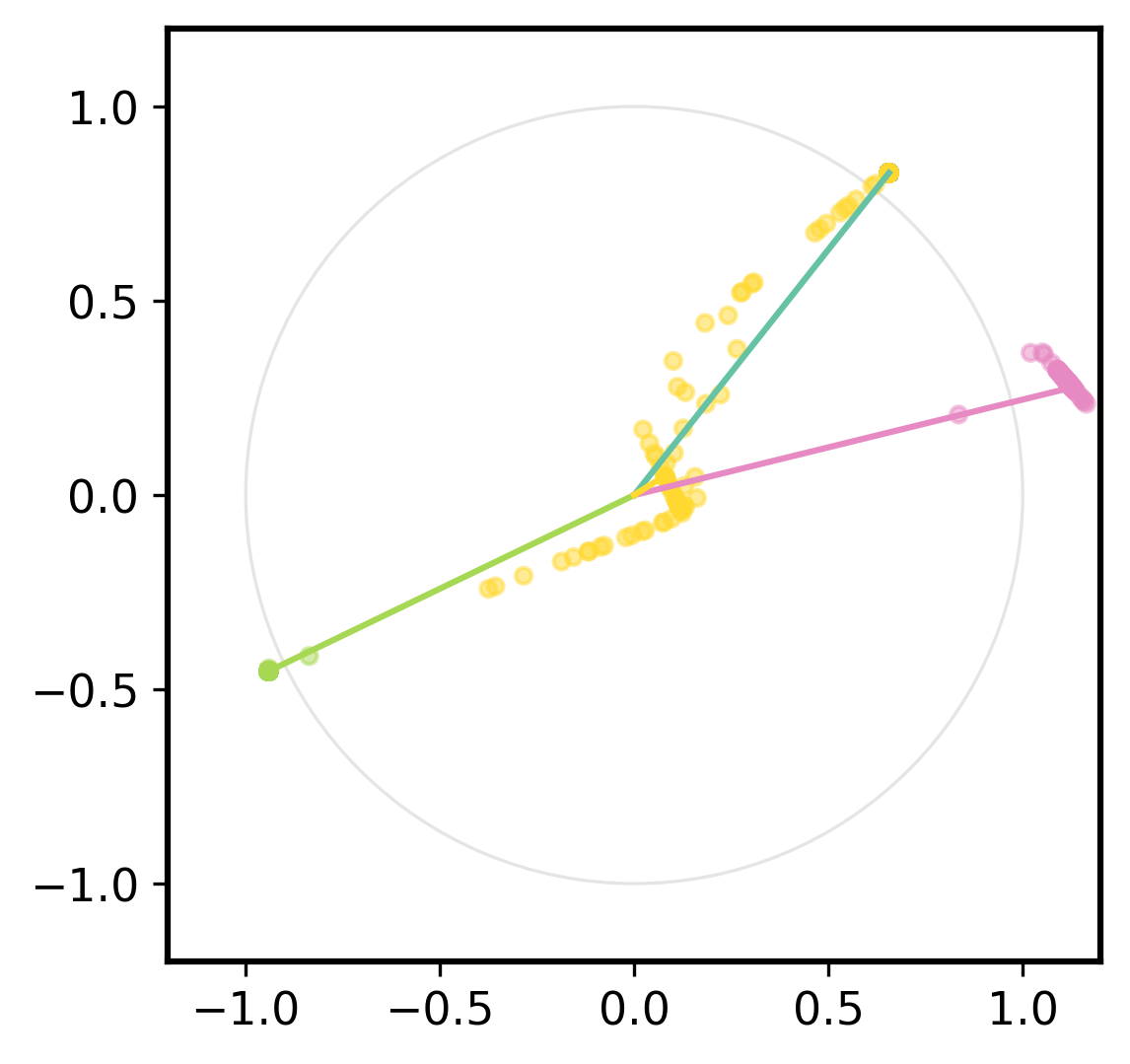}
        \caption{Client 0 (1)} \label{fig:client0_1}
    \end{subfigure}
    \hfill 
    \begin{subfigure}[b]{0.1135\textwidth}
        \includegraphics[width=\textwidth]{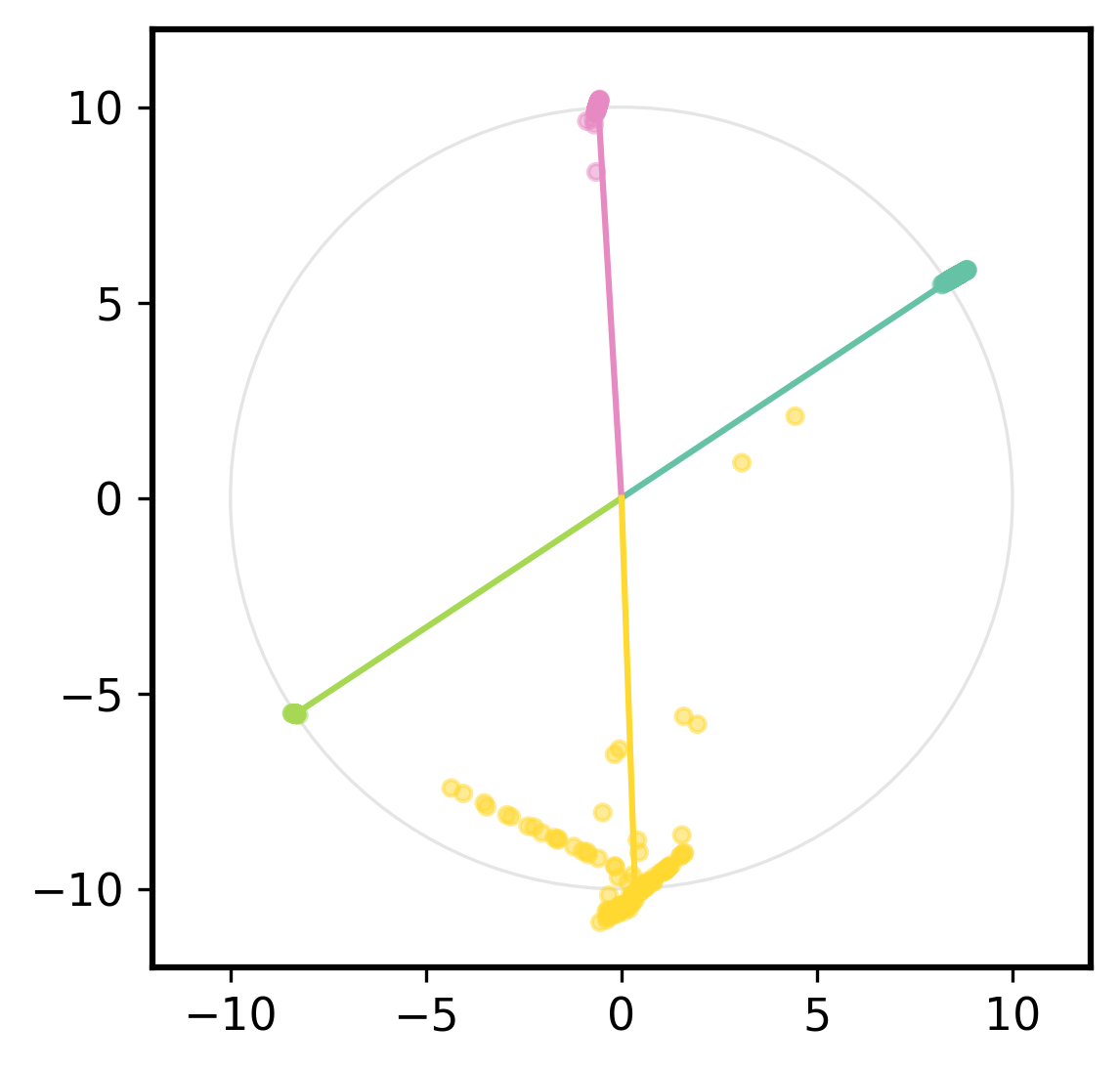}
        \caption{Client 0 (10)} \label{fig:client0_10}
    \end{subfigure}
    \hfill
    \begin{subfigure}[b]{0.115\textwidth}
        \includegraphics[width=\textwidth]{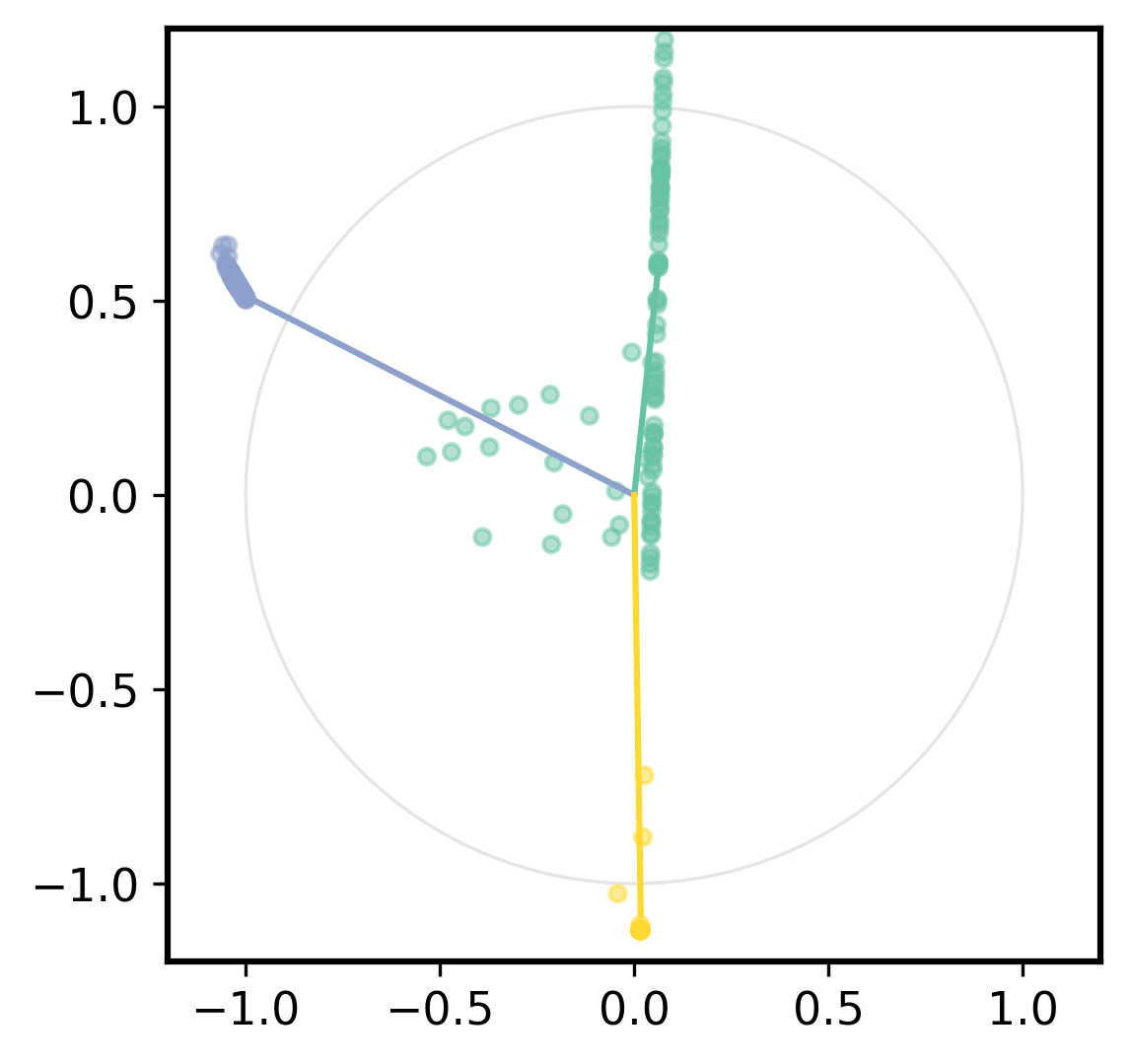}
        \caption{Client 1 (1)} 
    \end{subfigure}
    \hfill 
    \begin{subfigure}[b]{0.1135\textwidth}
        \includegraphics[width=\textwidth]{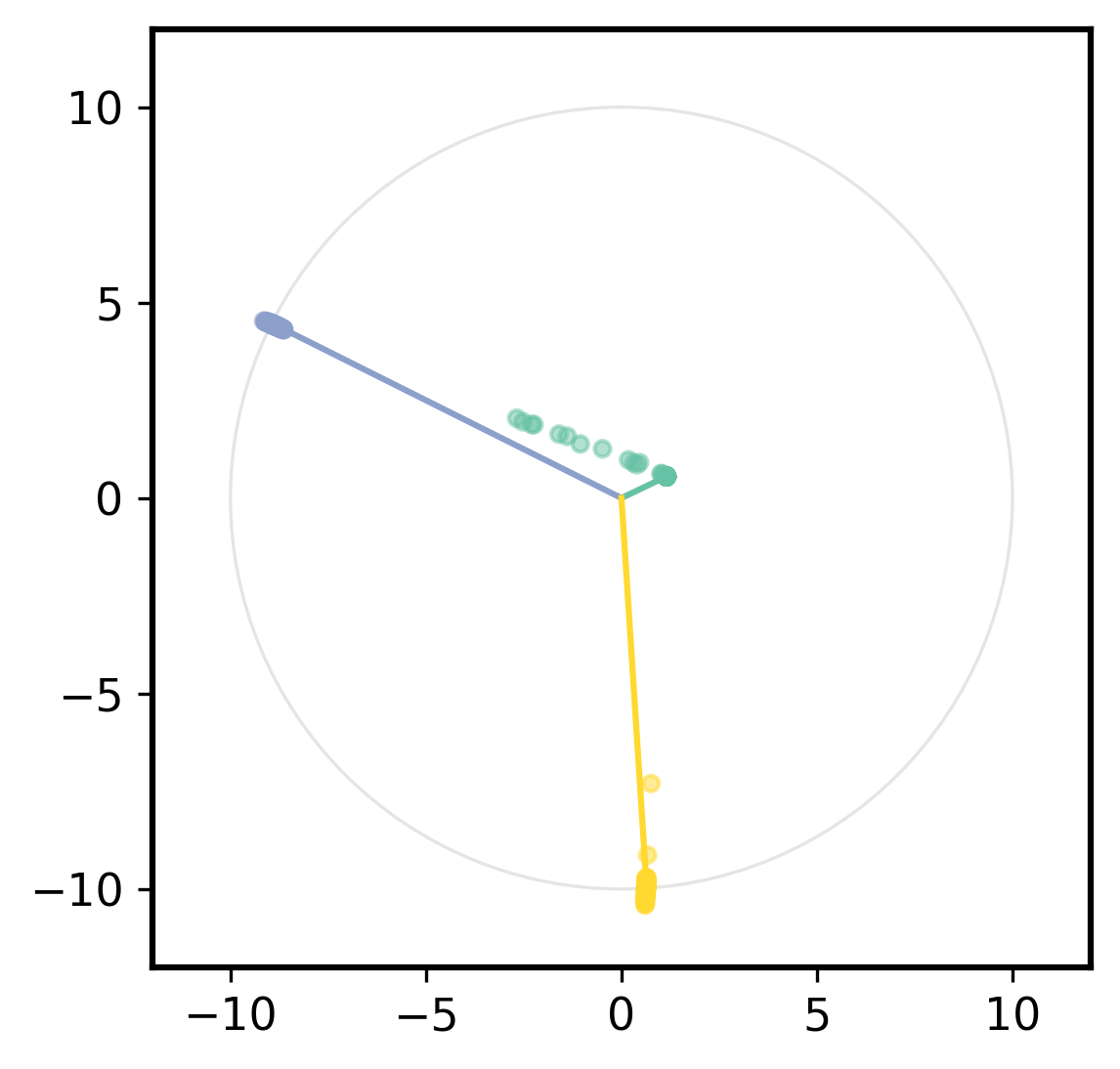}
        \caption{Client 1 (10)} 
    \end{subfigure}

    \begin{subfigure}[b]{0.115\textwidth}
        \includegraphics[width=\textwidth]{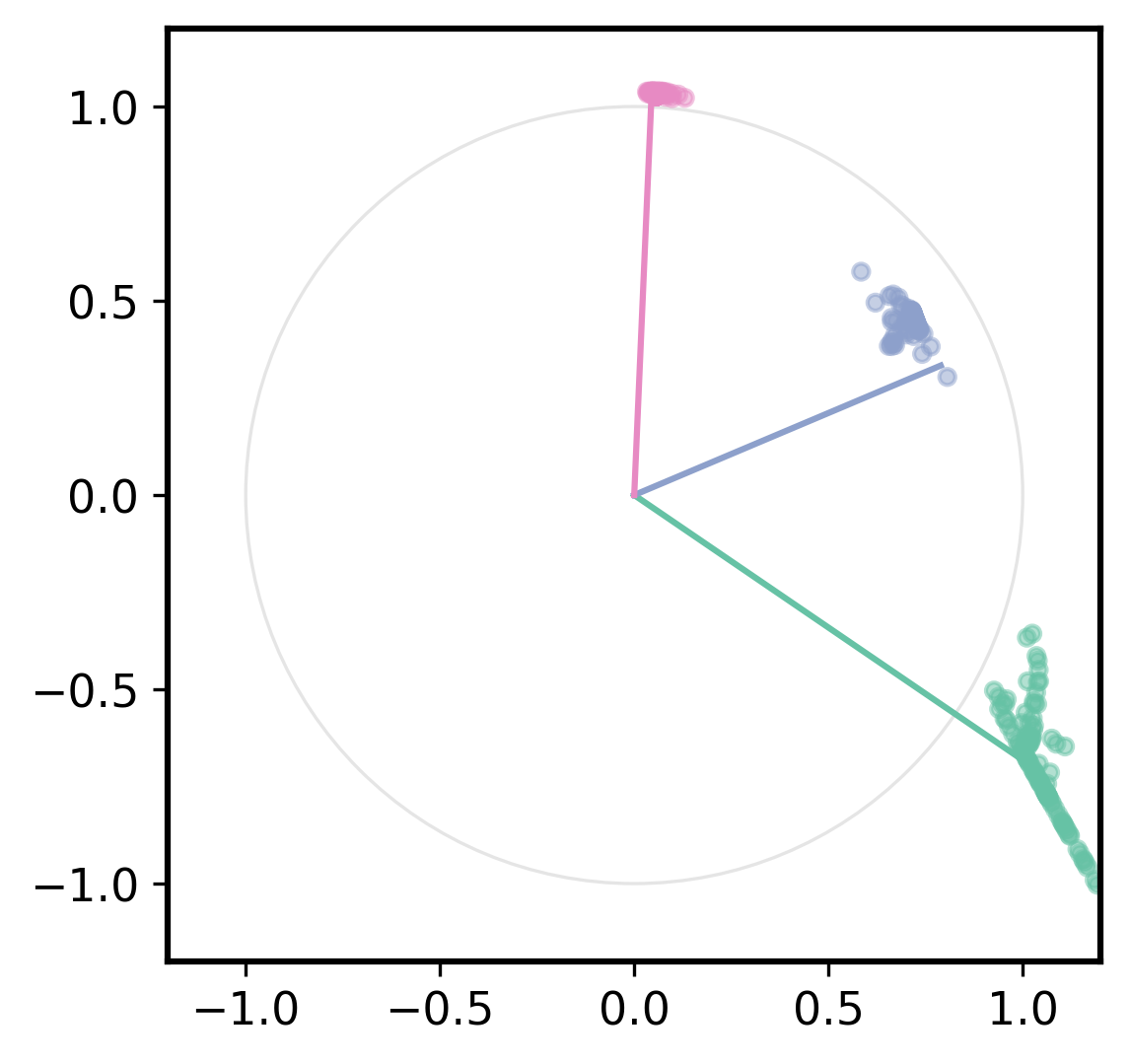}
        \caption{Client 2 (1)} 
    \end{subfigure}
    \hfill 
    \begin{subfigure}[b]{0.1135\textwidth}
        \includegraphics[width=\textwidth]{figures/spiral/client/a_visualization_ProtoNorm_lambda_0.5_csf_10.0_Client_ID_2.png}
        \caption{Client 2 (10)} 
    \end{subfigure}
    \hfill
    \begin{subfigure}[b]{0.115\textwidth}
        \includegraphics[width=\textwidth]{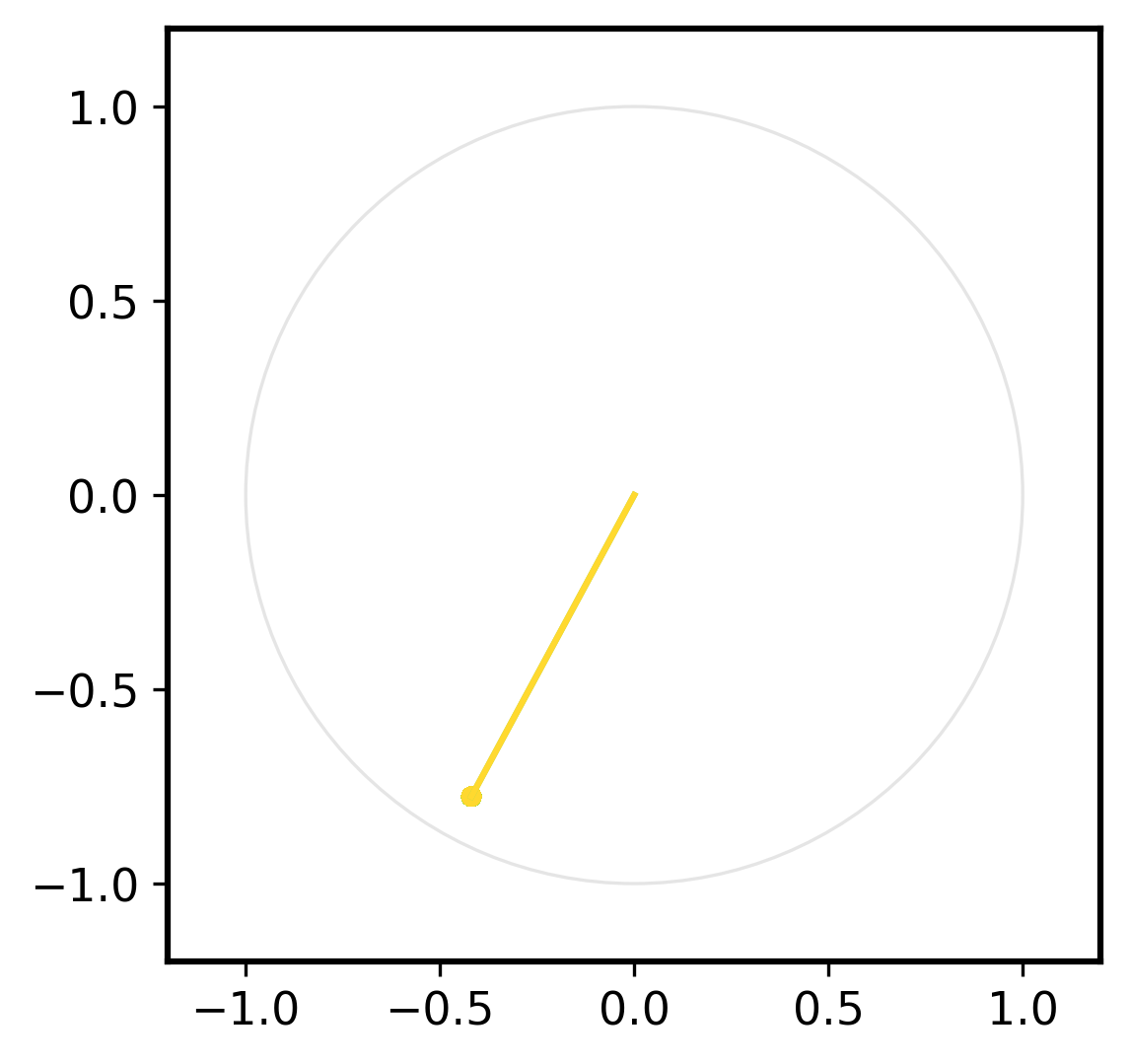}
        \caption{Client 3 (1)} 
    \end{subfigure}
    \hfill 
    \begin{subfigure}[b]{0.1135\textwidth}
        \includegraphics[width=\textwidth]{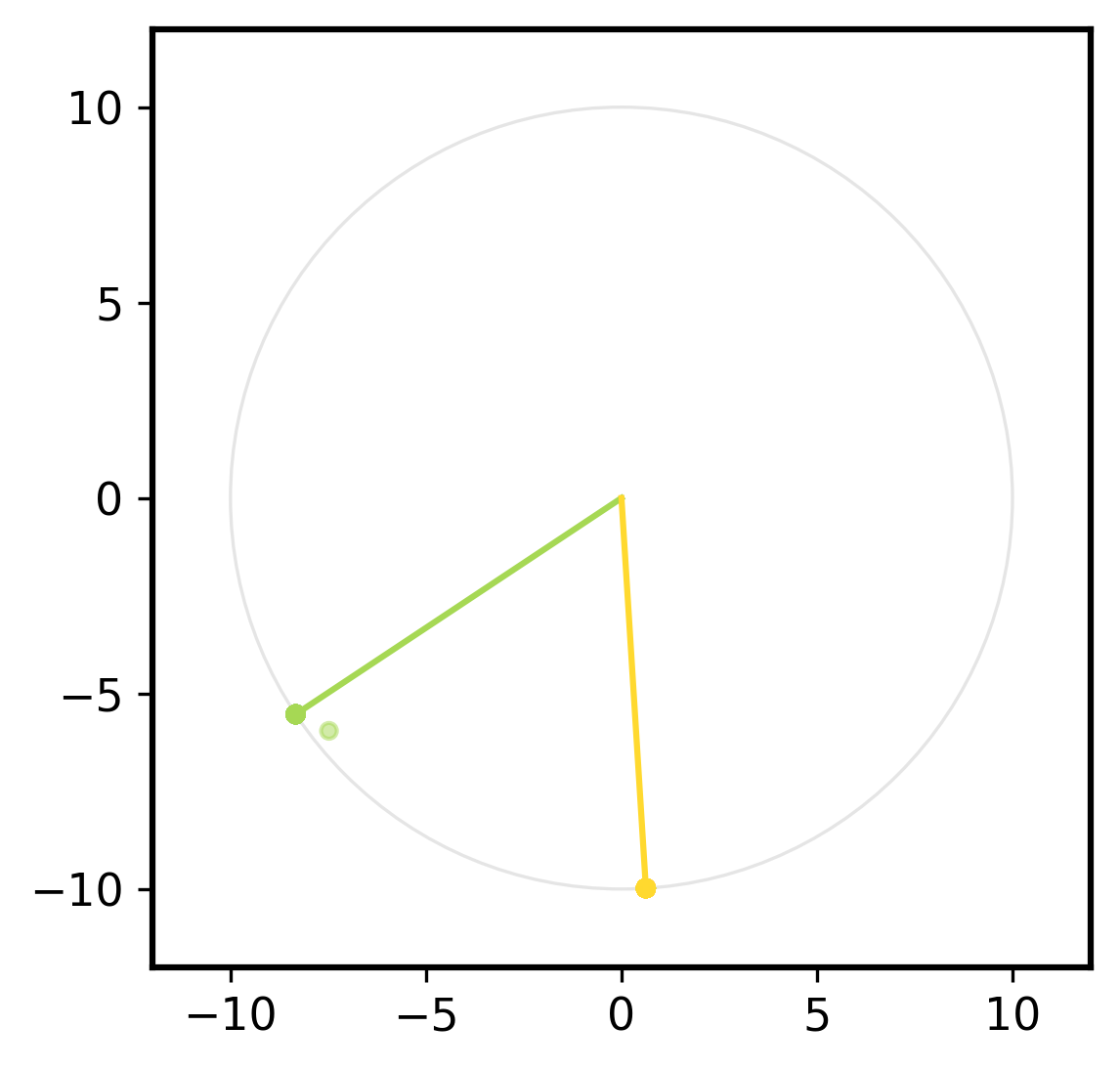}
        \caption{Client 3 (10)} 
    \end{subfigure}

    \begin{subfigure}[b]{0.115\textwidth}
        \includegraphics[width=\textwidth]{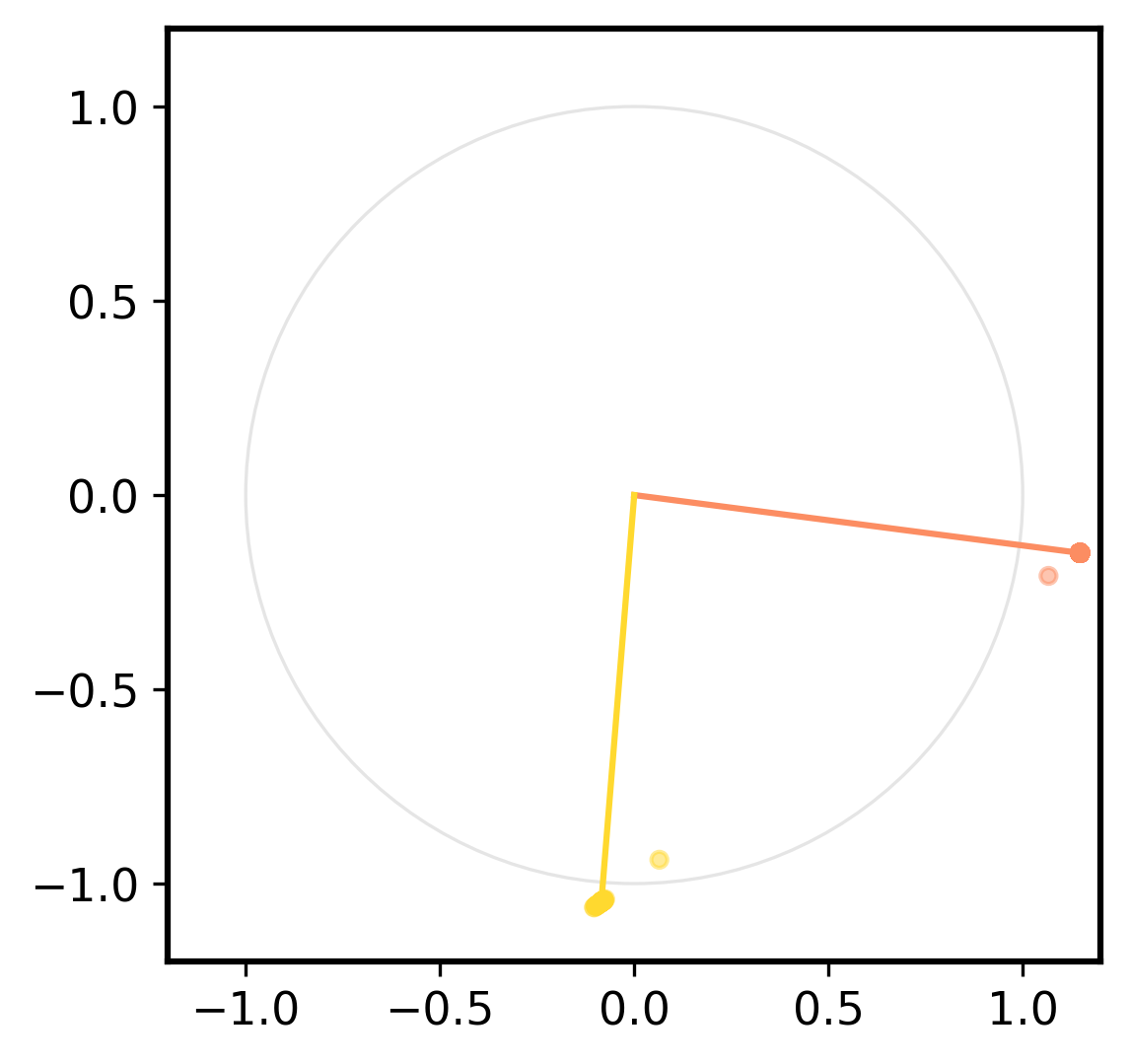}
        \caption{Client 4 (1)} 
    \end{subfigure}
    \hfill 
    \begin{subfigure}[b]{0.1135\textwidth}
        \includegraphics[width=\textwidth]{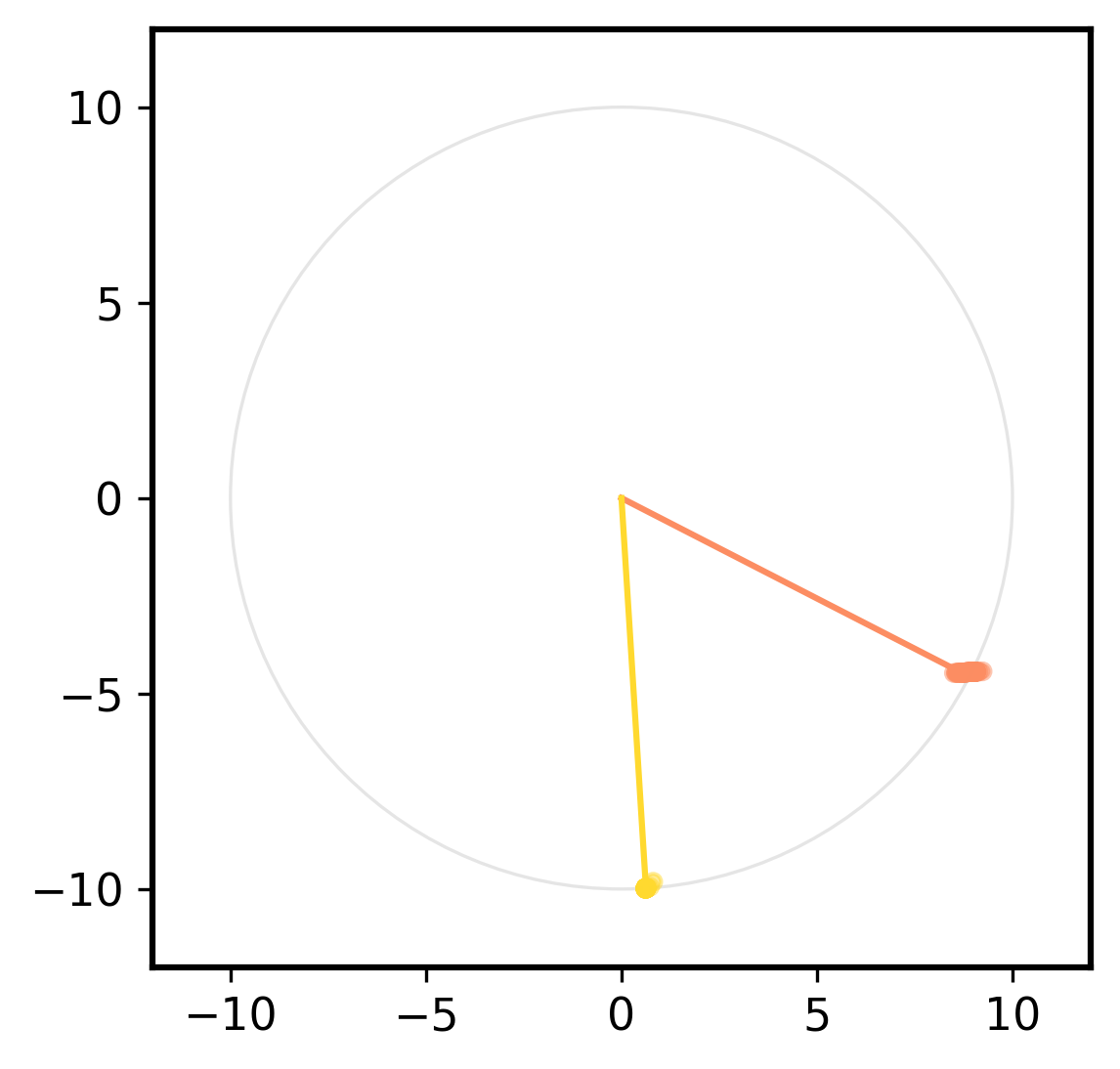}
        \caption{Client 4 (10)} 
    \end{subfigure}
    \hfill
    \begin{subfigure}[b]{0.115\textwidth}
        \includegraphics[width=\textwidth]{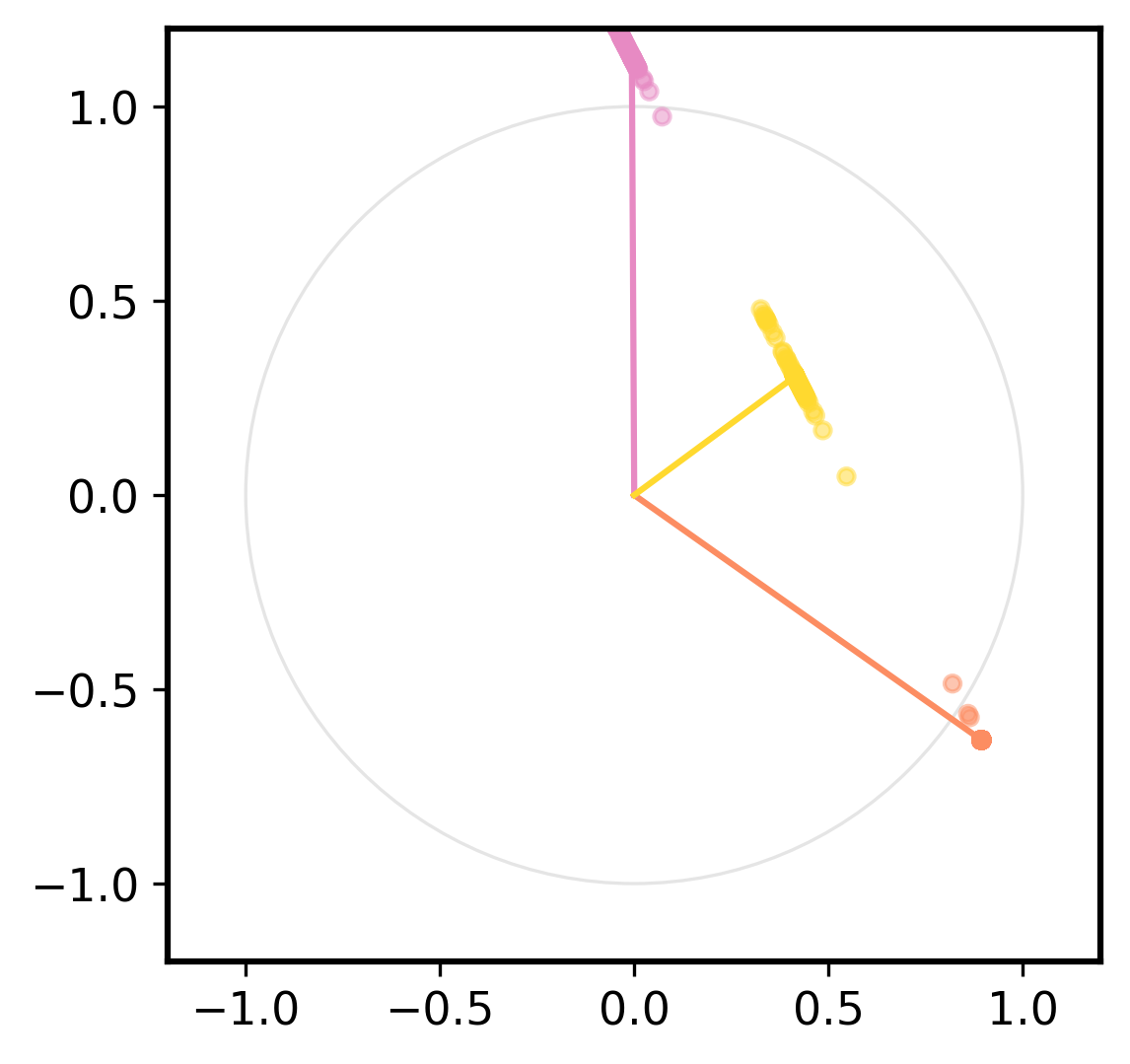}
        \caption{Client 5 (1)} 
    \end{subfigure}
    \hfill 
    \begin{subfigure}[b]{0.1135\textwidth}
        \includegraphics[width=\textwidth]{figures/spiral/client/a_visualization_ProtoNorm_lambda_0.5_csf_10.0_Client_ID_5.png}
        \caption{Client 5 (10)} 
    \end{subfigure}

    \begin{subfigure}[b]{0.115\textwidth}
        \includegraphics[width=\textwidth]{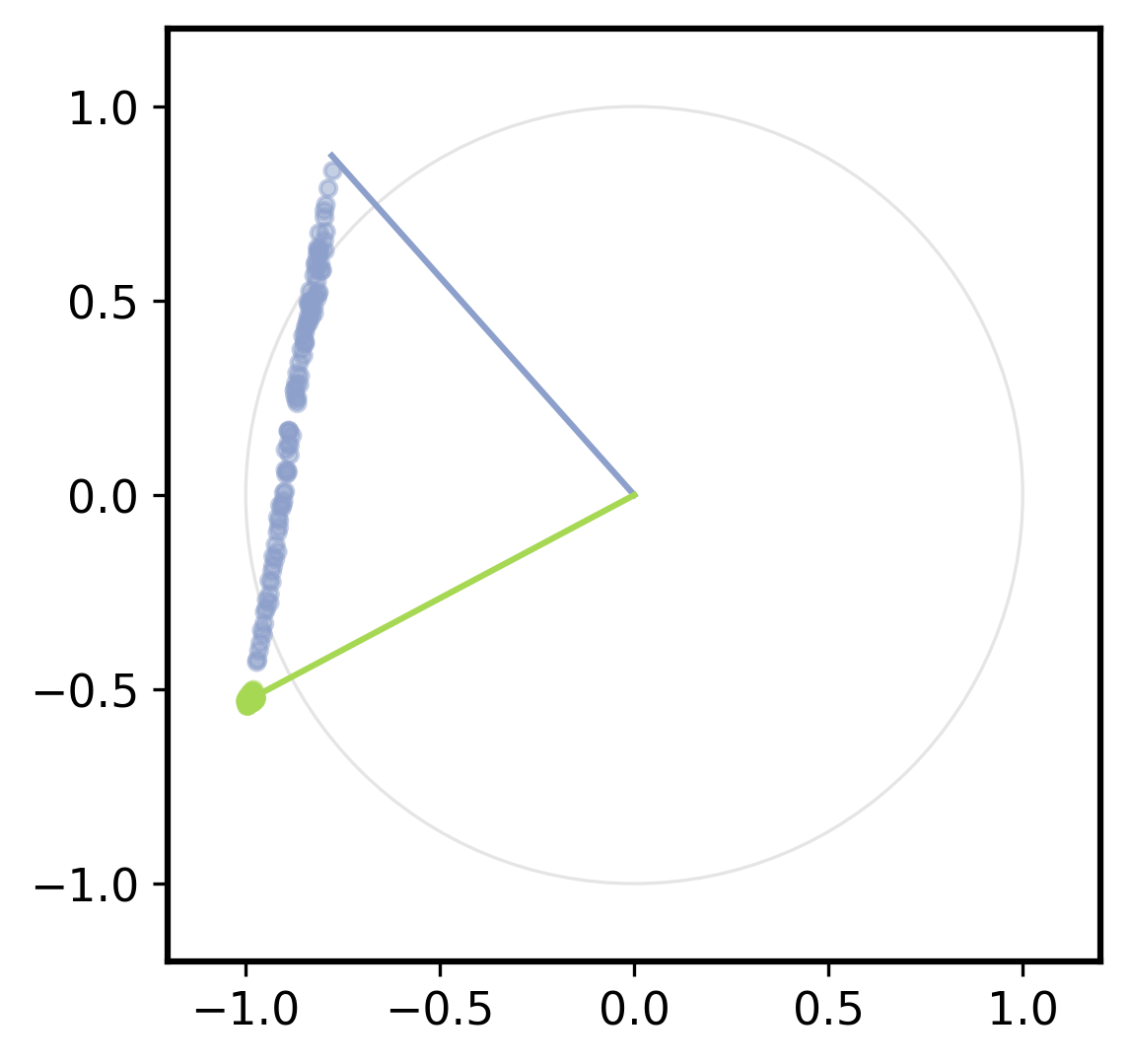}
        \caption{Client 6 (1)} 
    \end{subfigure}
    \hfill 
    \begin{subfigure}[b]{0.1135\textwidth}
        \includegraphics[width=\textwidth]{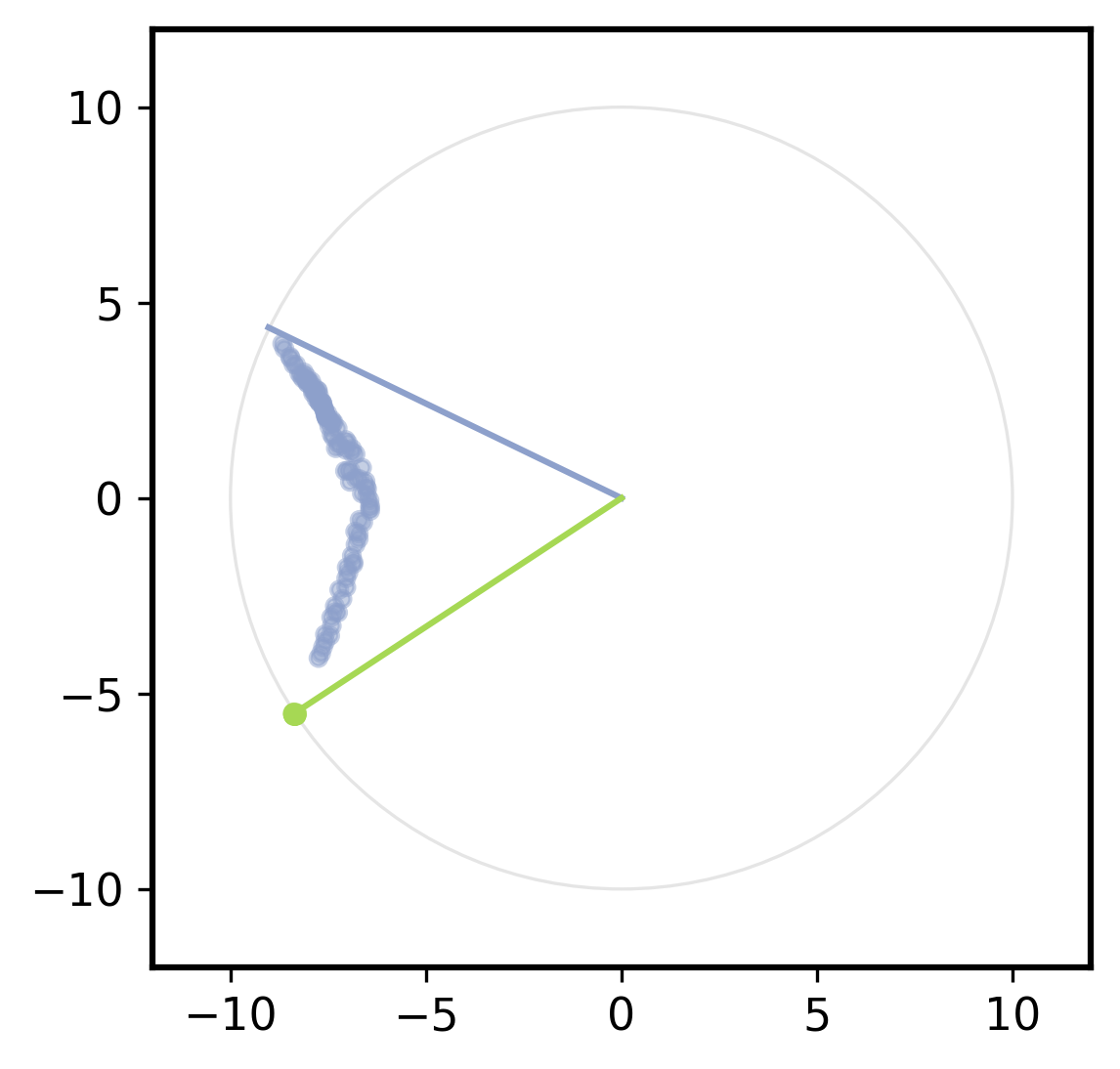}
        \caption{Client 6 (10)} 
    \end{subfigure}
    \hfill
    \begin{subfigure}[b]{0.115\textwidth}
        \includegraphics[width=\textwidth]{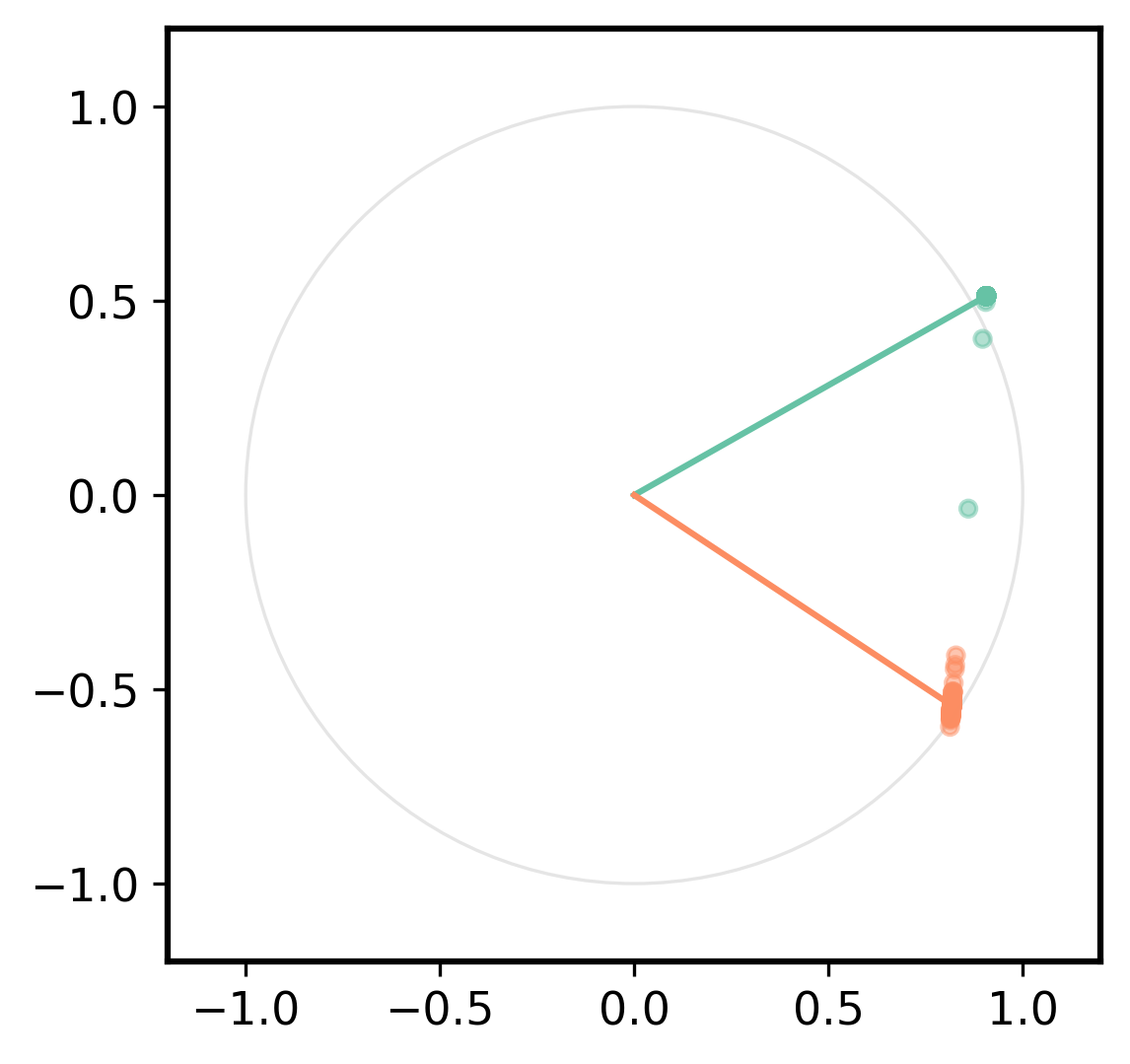}
        \caption{Client 7 (1)} 
    \end{subfigure}
    \hfill 
    \begin{subfigure}[b]{0.1135\textwidth}
        \includegraphics[width=\textwidth]{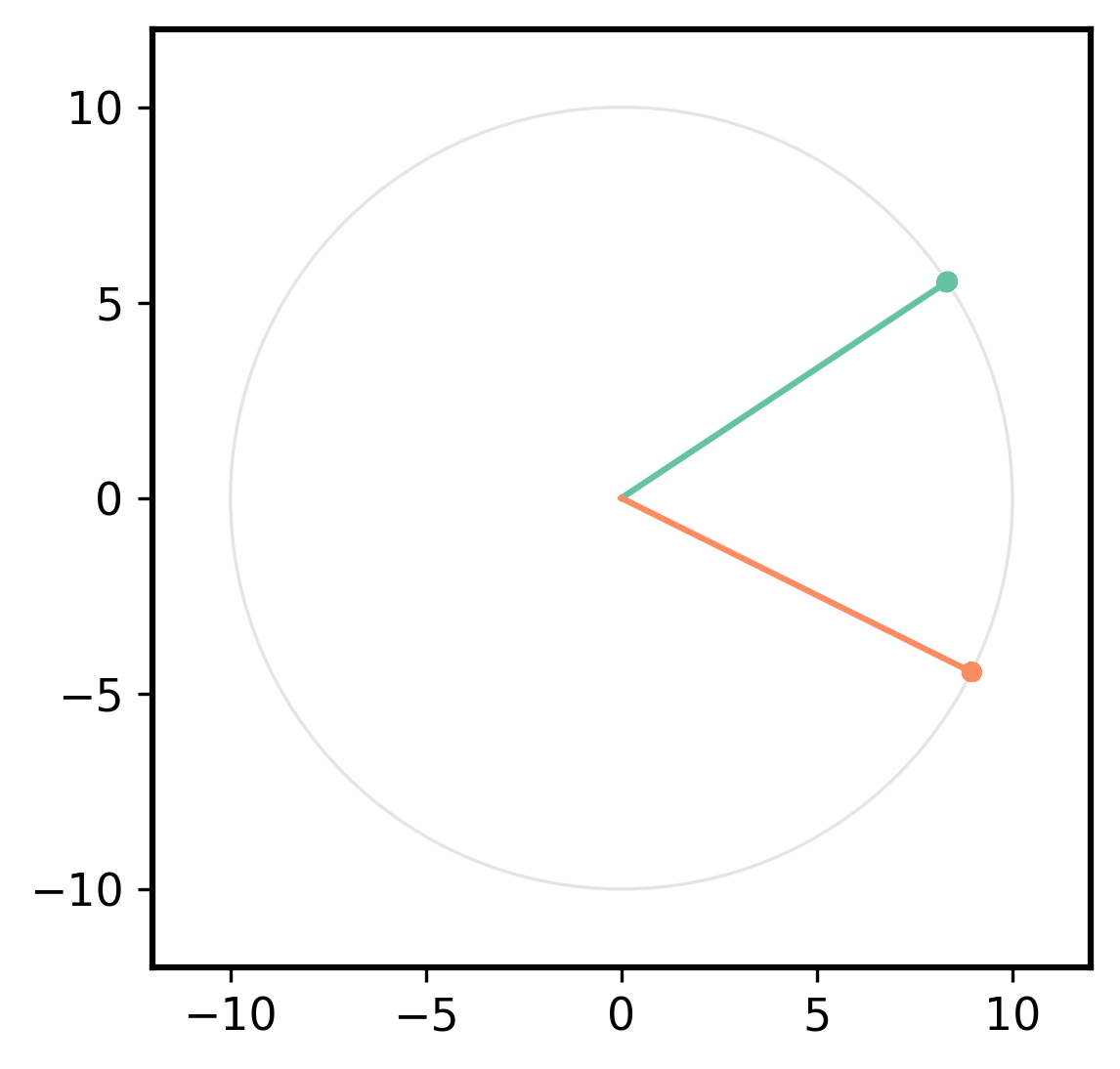}
        \caption{Client 7 (10)} 
    \end{subfigure}

    \begin{subfigure}[b]{0.115\textwidth}
        \includegraphics[width=\textwidth]{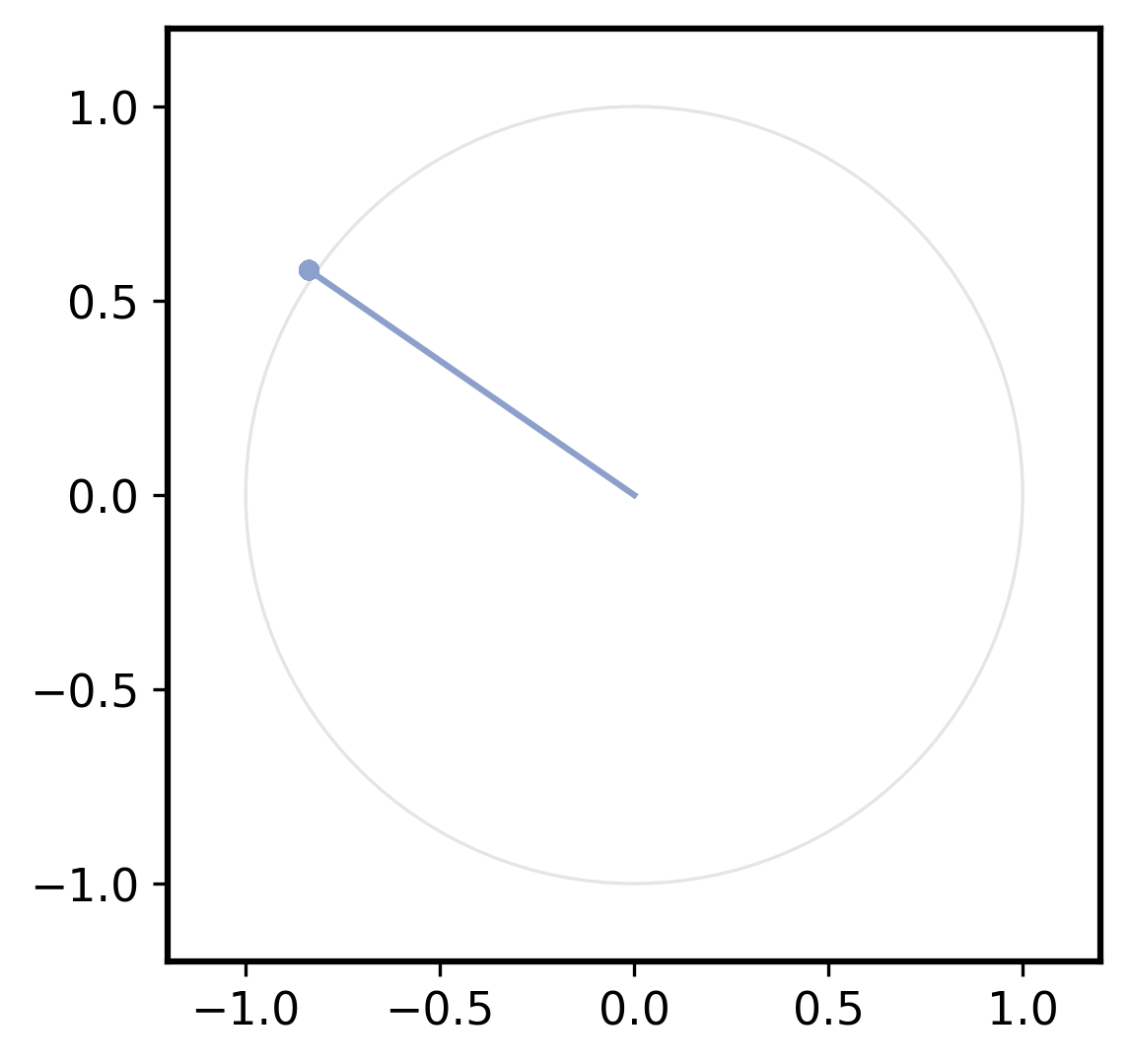}
        \caption{Client 8 (1)} 
    \end{subfigure}
    \hfill 
    \begin{subfigure}[b]{0.1135\textwidth}
        \includegraphics[width=\textwidth]{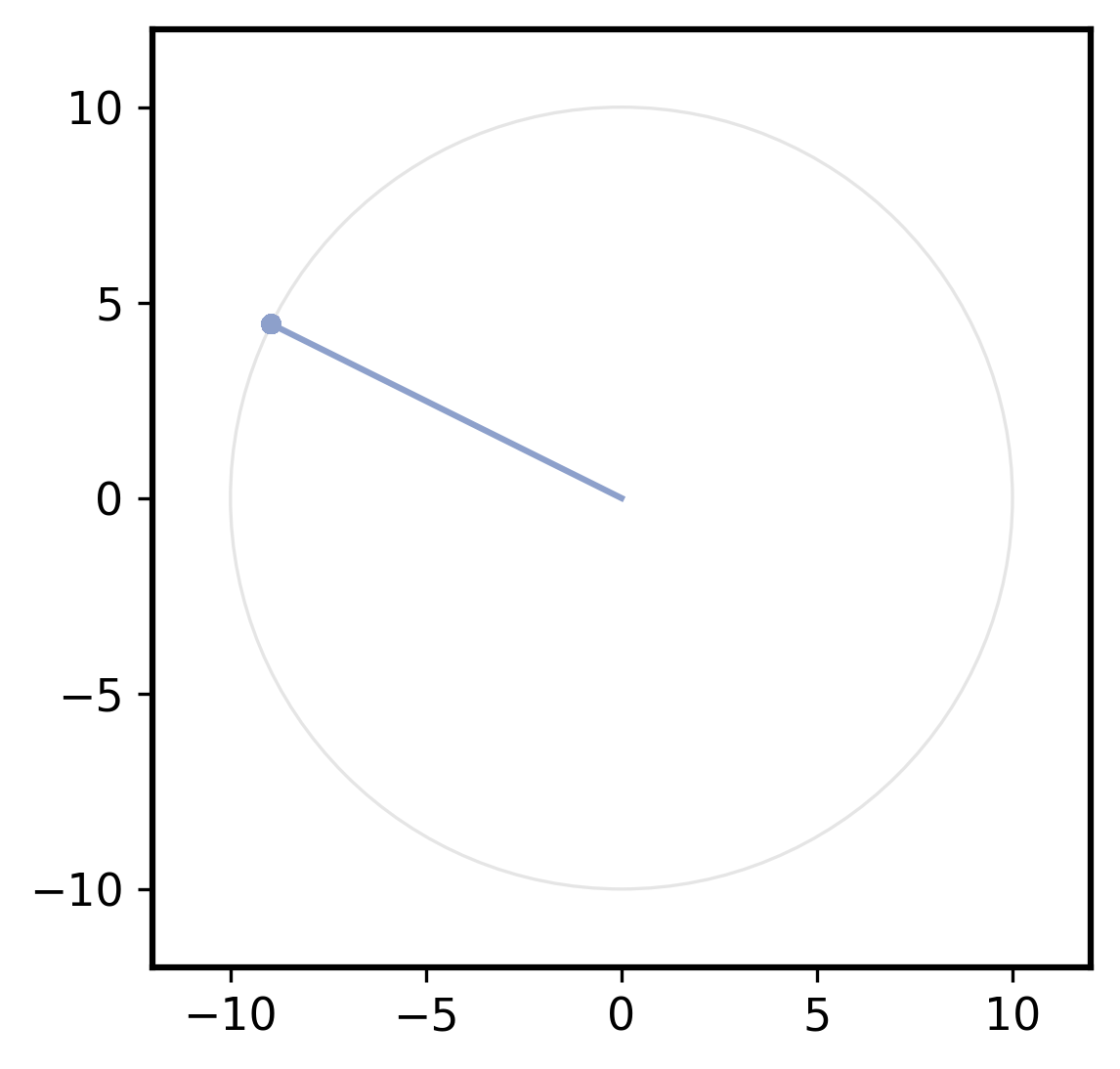}
        \caption{Client 8 (10)} 
    \end{subfigure}
    \hfill
    \begin{subfigure}[b]{0.115\textwidth}
        \includegraphics[width=\textwidth]{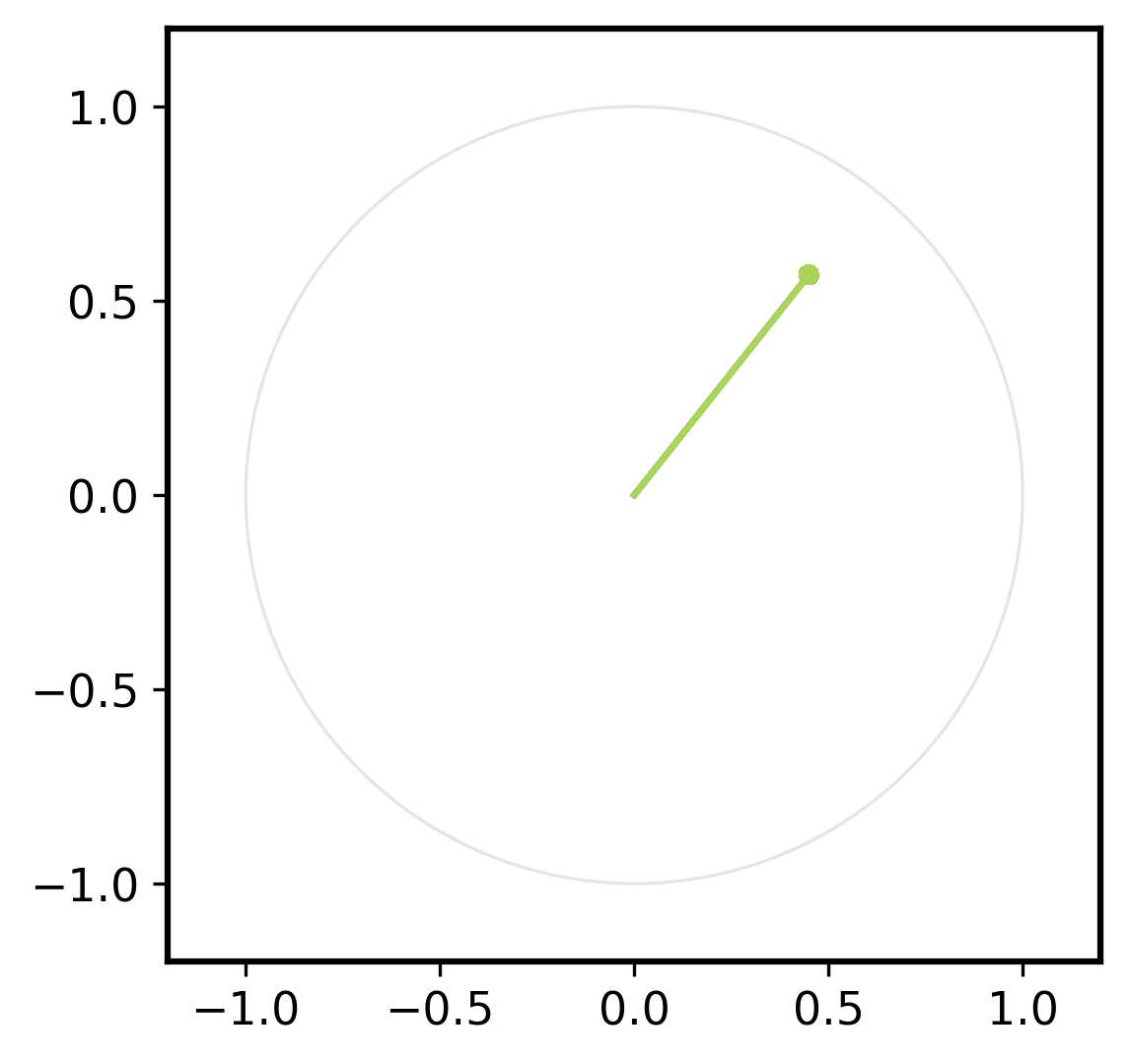}
        \caption{Client 9 (1)} 
    \end{subfigure}
    \hfill 
    \begin{subfigure}[b]{0.1135\textwidth}
        \includegraphics[width=\textwidth]{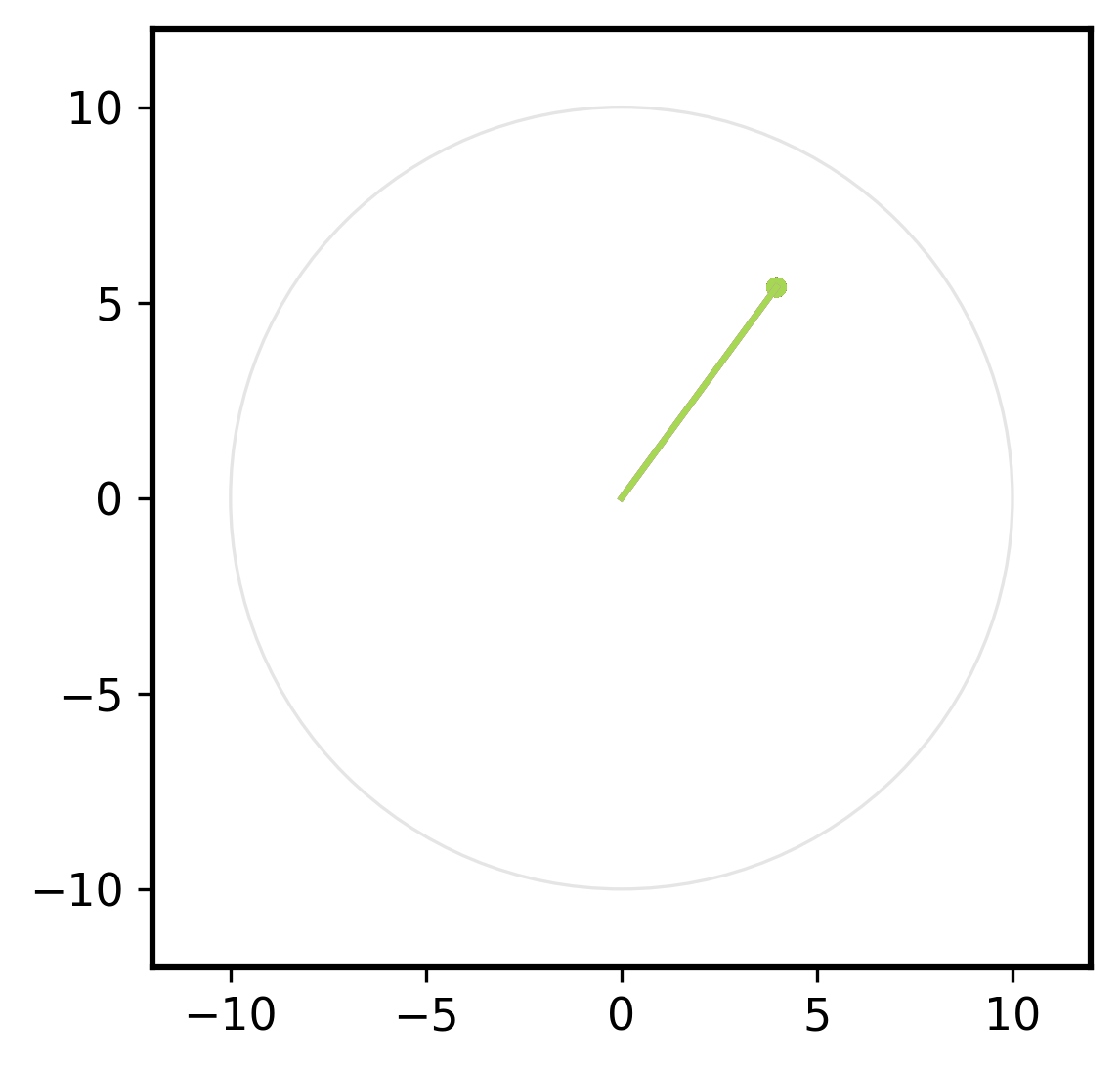}
        \caption{Client 9 (10)} 
    \end{subfigure}
    \caption{Local prototypes and learned representations of clients. Number in parenthesis indicates $\gamma$.}
    \label{fig:localproto_representation}
\end{figure}
\noindent 
We illustrate clients' local prototypes and learned representations of test samples for ProtoNorm. Results demonstrate that ProtoNorm with $\gamma=10$ consistently achieves better prototype separation than its non-scaled counterpart. For instance, yellow label samples (class 5) in Figure~\ref{fig:client0_1} and ~\ref{fig:client0_10} are difficult to classify due to their scarcity in the training set of client 0, as illustrated in Figure~\ref{fig:toy_dataset}. ProtoNorm with large $\gamma$ (Figure~\ref{fig:client0_10}) handles this class imbalance more effectively than small $\gamma$ (Figure~\ref{fig:client0_1}). Notably, even for clients with abundant samples and consequently less challenging tasks (clients 2 and 5), ProtoNorm with large $\gamma$ demonstrates superior performance.

\section*{C. Experimental Details}  
\subsection*{C.1 Hyperparameters} 
For baseline algorithms, we implement hyperparameters following recommendations in \cite{zhang2024fedtgp}. Table \ref{table:hp_settings} summarizes these configurations. Note that hyperparameter notations in Table \ref{table:hp_settings} are algorithm-specific and may not correspond to notations used elsewhere in this paper.
\begin{table}[bht]
    \centering
    {\fontsize{9}{11}\selectfont
    \begin{tabular}{lp{0.76\linewidth}}
    \toprule
    Method      & Hyperparameter settings \\
    \midrule
    \multirow{2}{*}{FML} & $\alpha$ (KD weight for local model) $= 0.5$ \\
                & $\beta$ (KD weight for meme model) $= 0.5$ \\
    \midrule
    \multirow{2}{*}{FedKD} & $T_{\text{start}}$ (energy threshold) $= 0.95$ \\
                & $T_{\text{end}}$ (energy threshold) $= 0.98$ \\
    \midrule
    FedDistill  & $\gamma$ (weight of logit regularizer) $= 1$ \\
    \midrule
    FedNH & $\rho$ \begin{footnotesize}(smoothing parameter)\end{footnotesize} $ = 0.9$ \\
   \midrule
   \multirow{2}{*}{FedUV} & $\mu$ (classifier variance regularizer) $ = 2.5$ \\
   & $\lambda$ (Hyperspherical uniformity regularizer) $ = 0.5$ \\
    \bottomrule
    \end{tabular}
    }
    \caption{Hyperparameter settings for the compared methods.}
    \label{table:hp_settings}
    % \vspace{-10pt}
\end{table}

\subsection*{C.2 Experimental Environment} 
To ensure reproducibility, we conducted all experiments in a standardized environment with the following specifications:
\begin{itemize}
    \item Framework: PyTorch 2.4
    \item Hardware:
    \begin{itemize}
        \item CPUs: 2 Intel Xeon Gold 6240R (96 cores total)
        \item Memory: 256GB
        \item GPUs: Two NVIDIA RTX A6000
    \end{itemize}
    \item Operating System: Ubuntu 22.04 LTS
\end{itemize}
This consistent configuration enabled reliable performance evaluation across all methods. The implementation code is available in the supplementary materials.

\subsection*{C.3. Data Distributions of Practical Settings} 
\begin{figure}[htb]
    \centering
    \begin{subfigure}[b]{0.20\textwidth}
        \includegraphics[width=\textwidth]{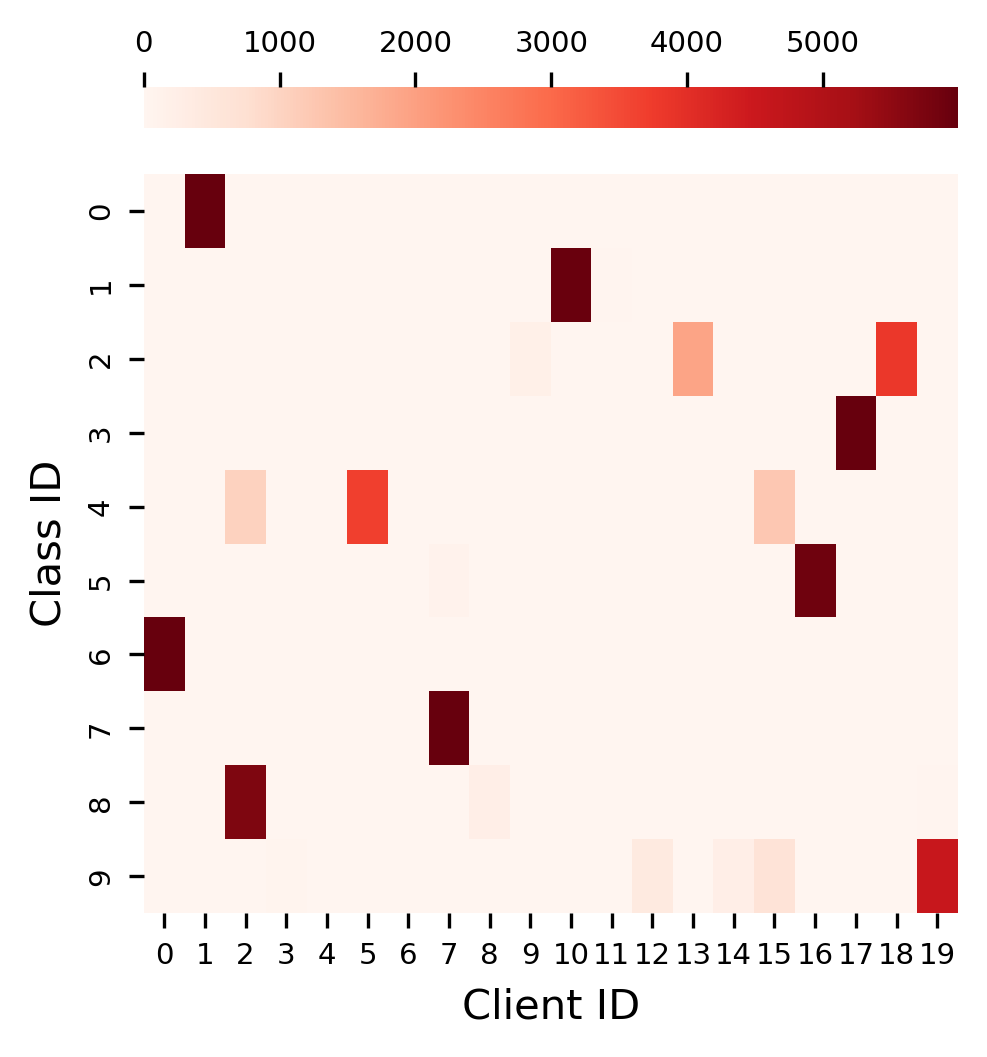}
        \caption{$\alpha=0.01$} 
    \end{subfigure}
    \quad 
    \begin{subfigure}[b]{0.20\textwidth}
        \includegraphics[width=\textwidth]{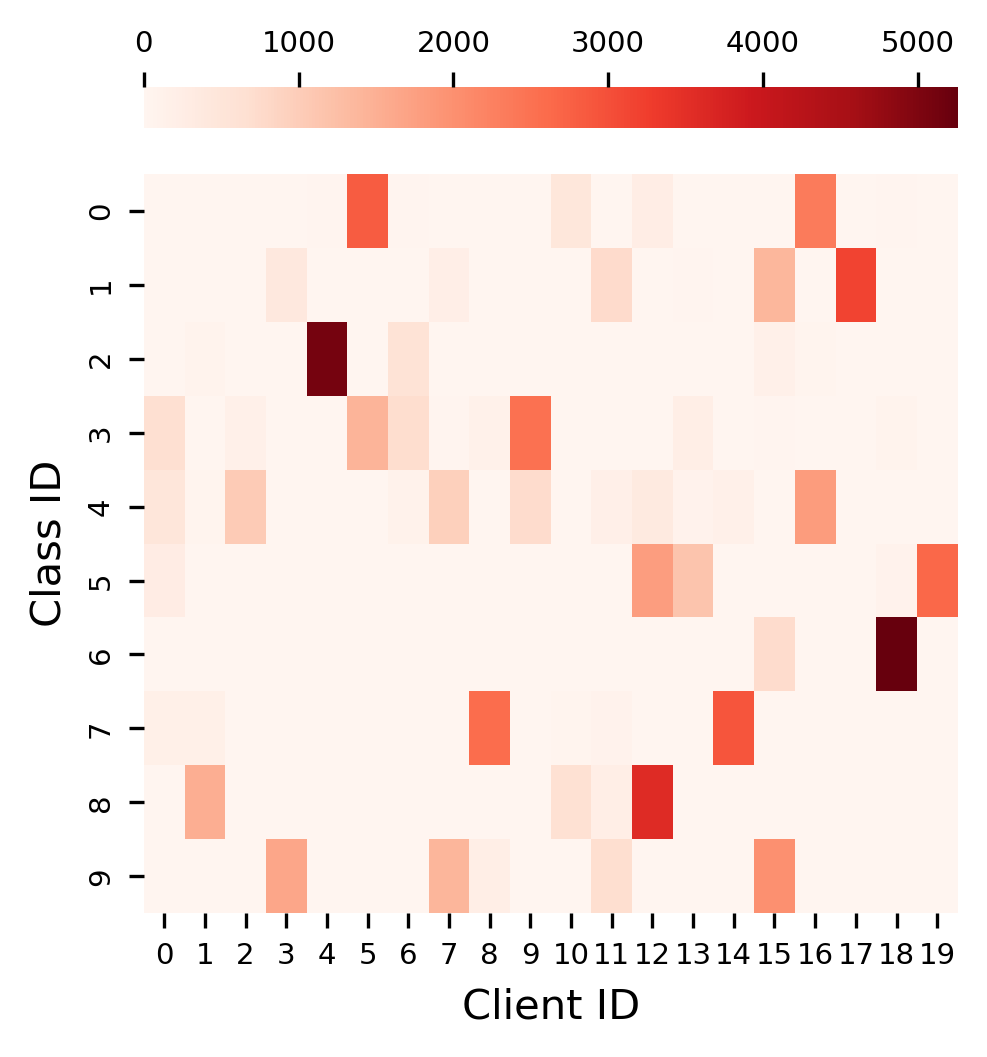}
        \caption{$\alpha=0.1$} 
    \end{subfigure}

    \begin{subfigure}[b]{0.20\textwidth}
        \includegraphics[width=\textwidth]{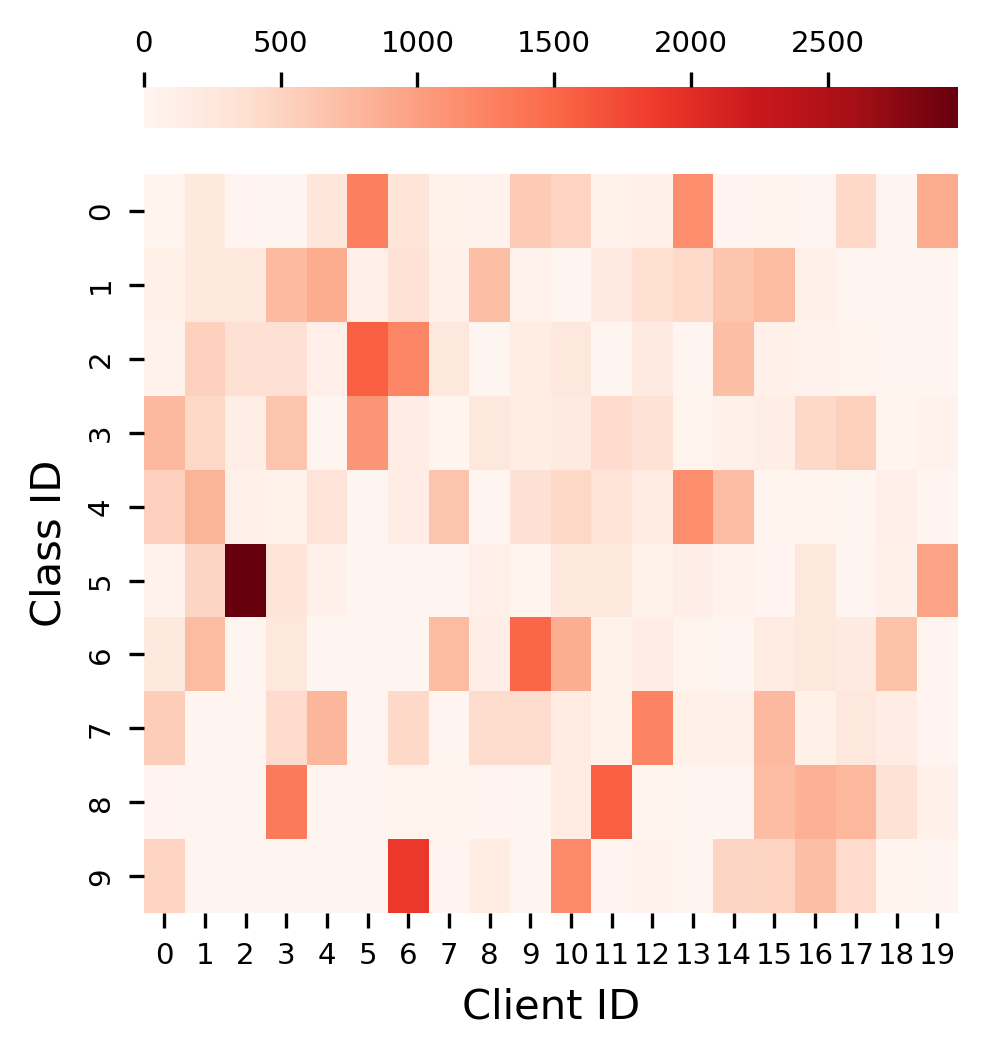}
        \caption{$\alpha=0.5$} 
    \end{subfigure}
    \quad 
    \begin{subfigure}[b]{0.20\textwidth}
        \includegraphics[width=\textwidth]{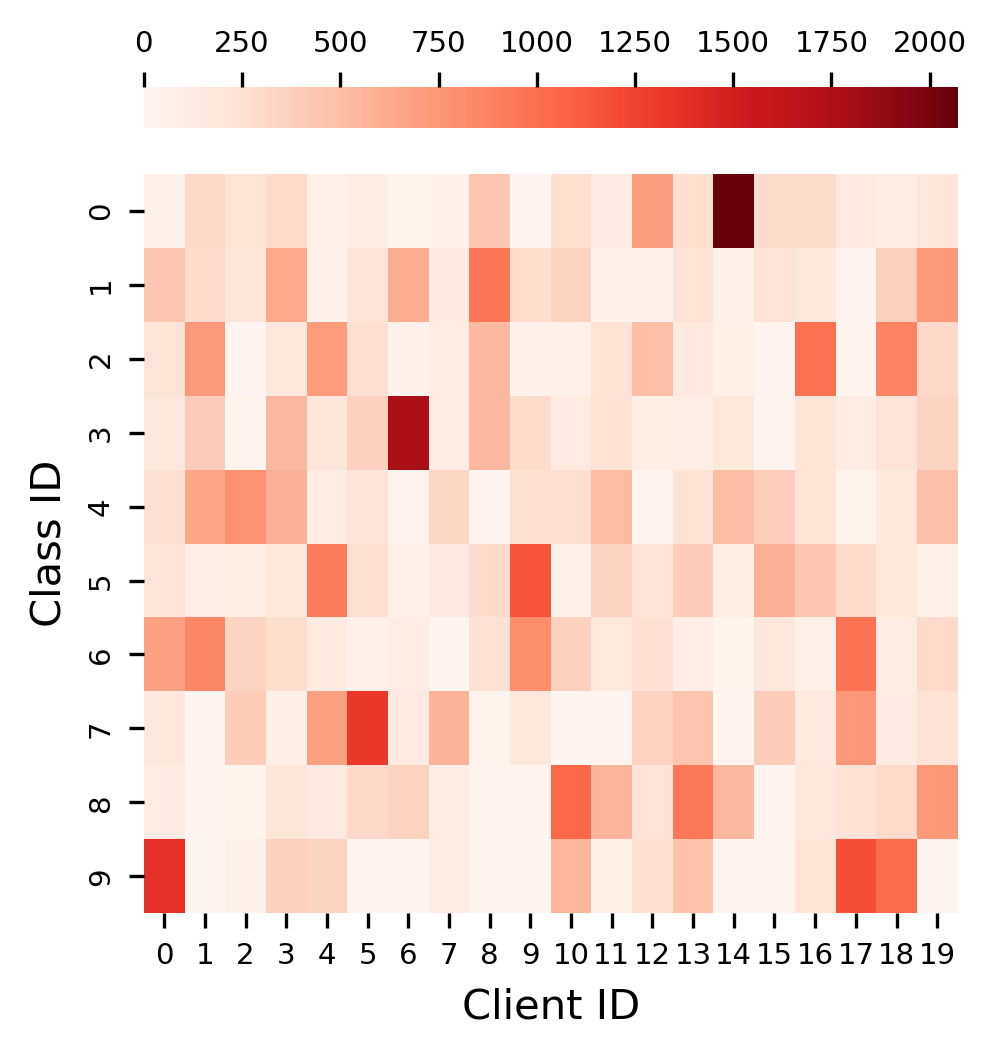}
        \caption{$\alpha=1.0$} 
    \end{subfigure}
    \caption{Data distributions for the CIFAR-10 practical setting.}
    \label{fig:distribution_cifar10}
    \vspace{-10pt}
\end{figure}

\begin{figure}[htb]
    \centering
    \begin{subfigure}[b]{0.20\textwidth}
        \includegraphics[width=\textwidth]{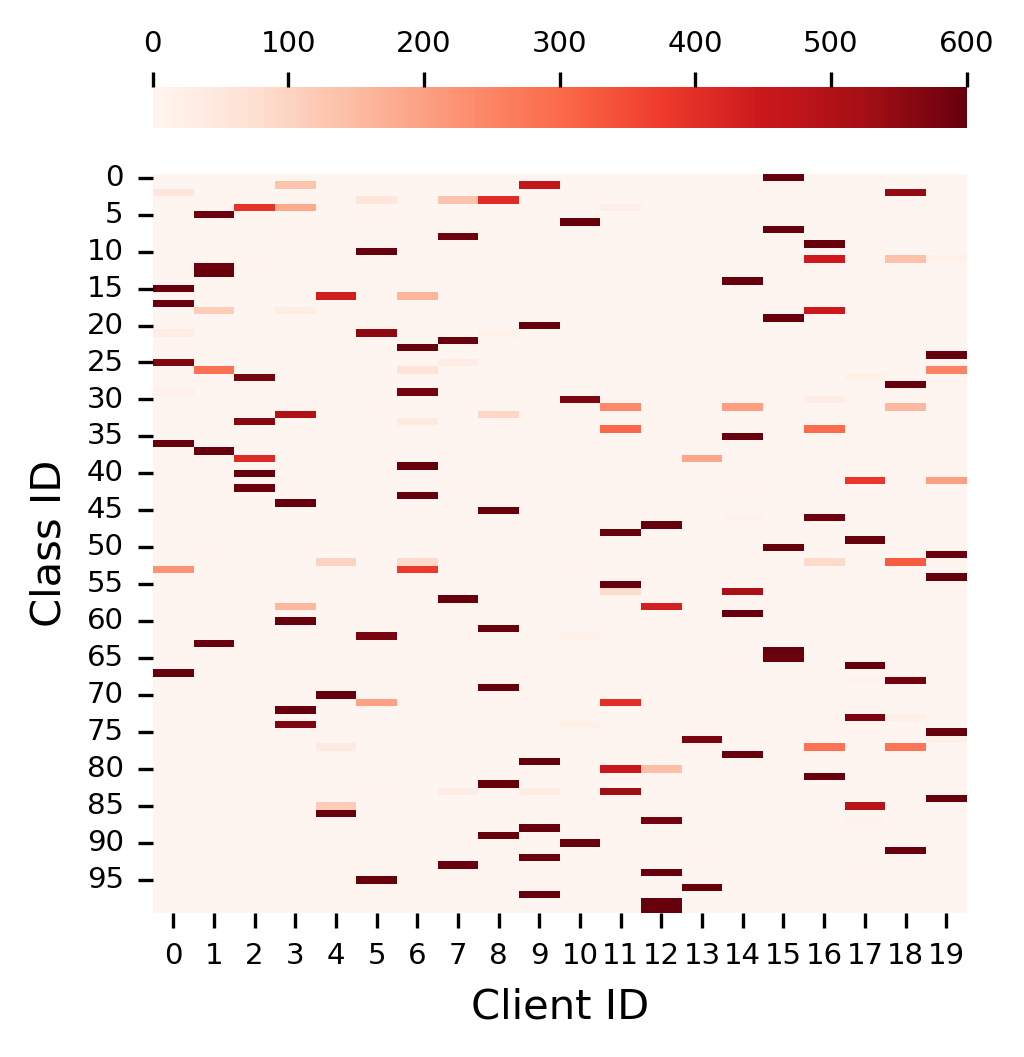}
        \caption{$\alpha=0.01$} 
    \end{subfigure}
    \quad 
    \begin{subfigure}[b]{0.20\textwidth}
        \includegraphics[width=\textwidth]{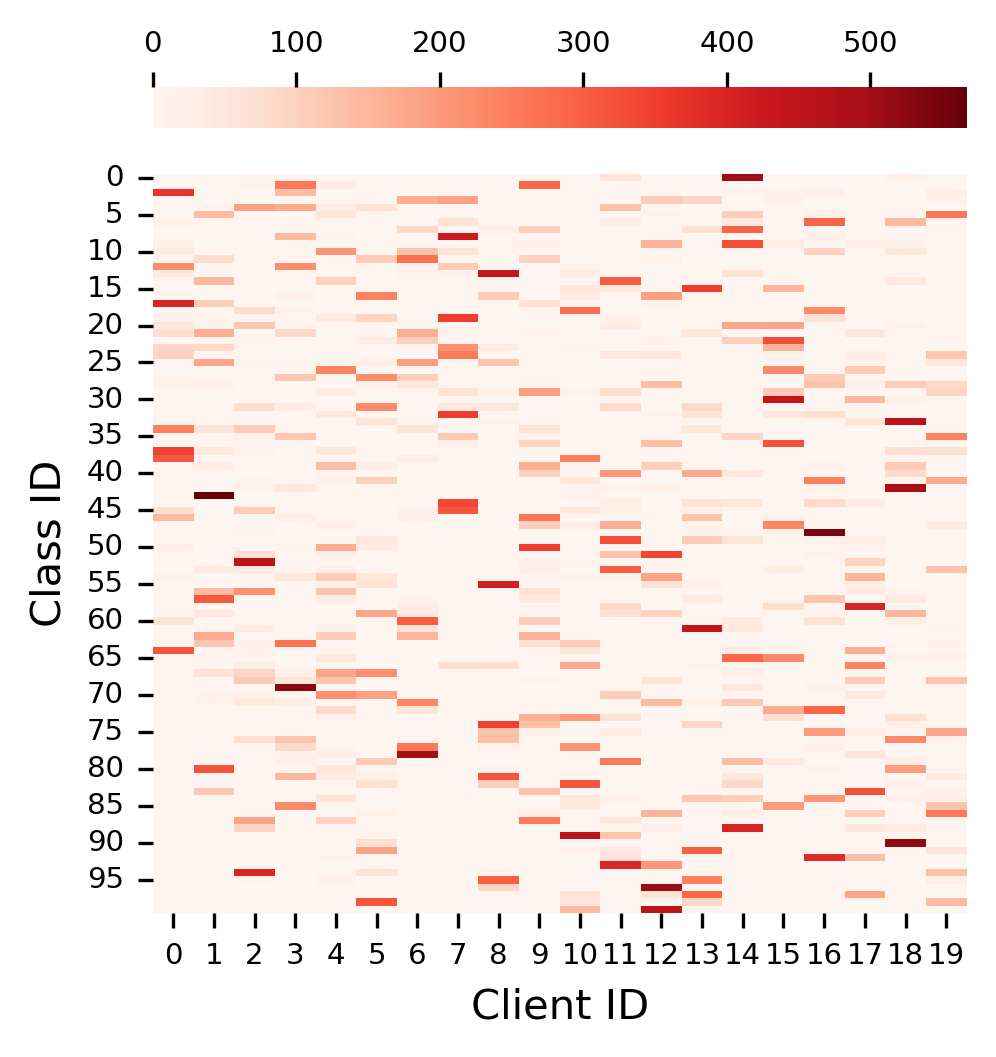}
        \caption{$\alpha=0.1$} 
    \end{subfigure}

    \begin{subfigure}[b]{0.20\textwidth}
        \includegraphics[width=\textwidth]{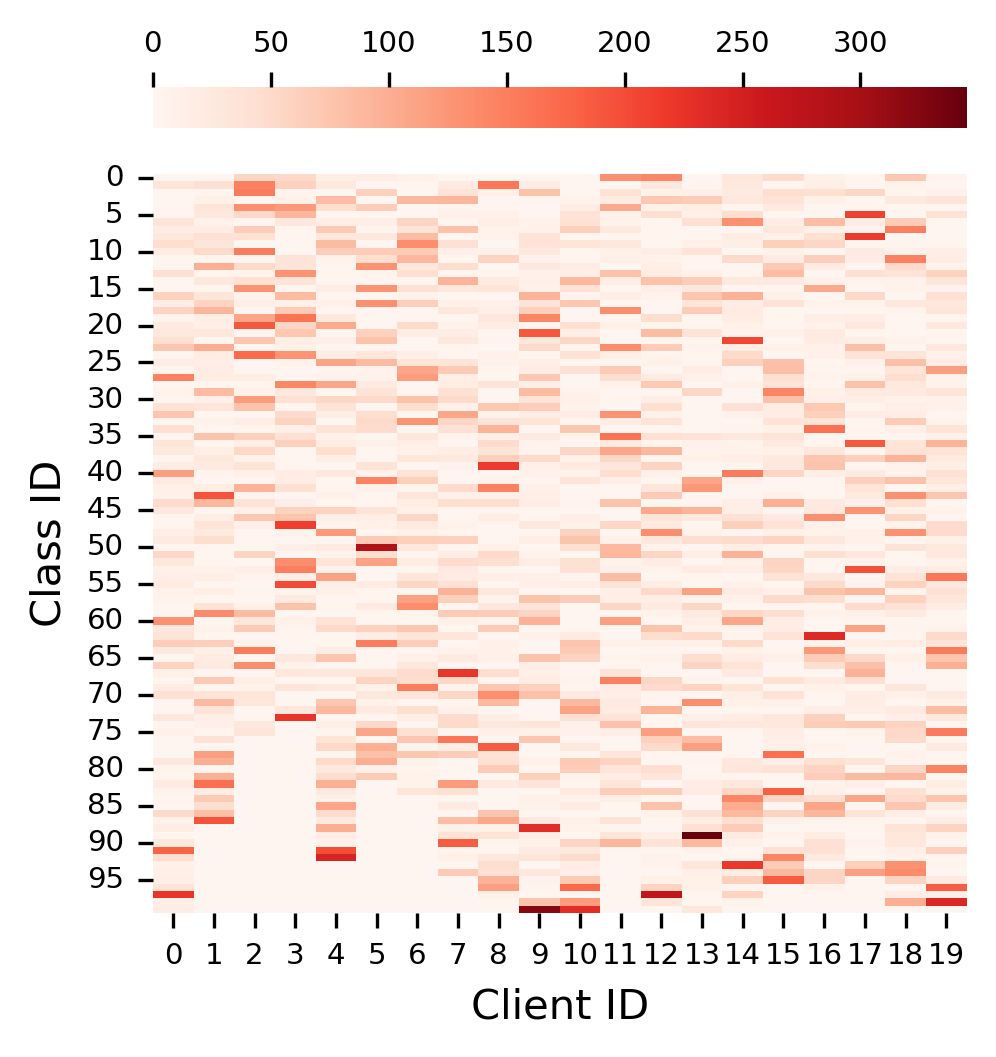}
        \caption{$\alpha=0.5$} 
    \end{subfigure}
    \quad 
    \begin{subfigure}[b]{0.20\textwidth}
        \includegraphics[width=\textwidth]{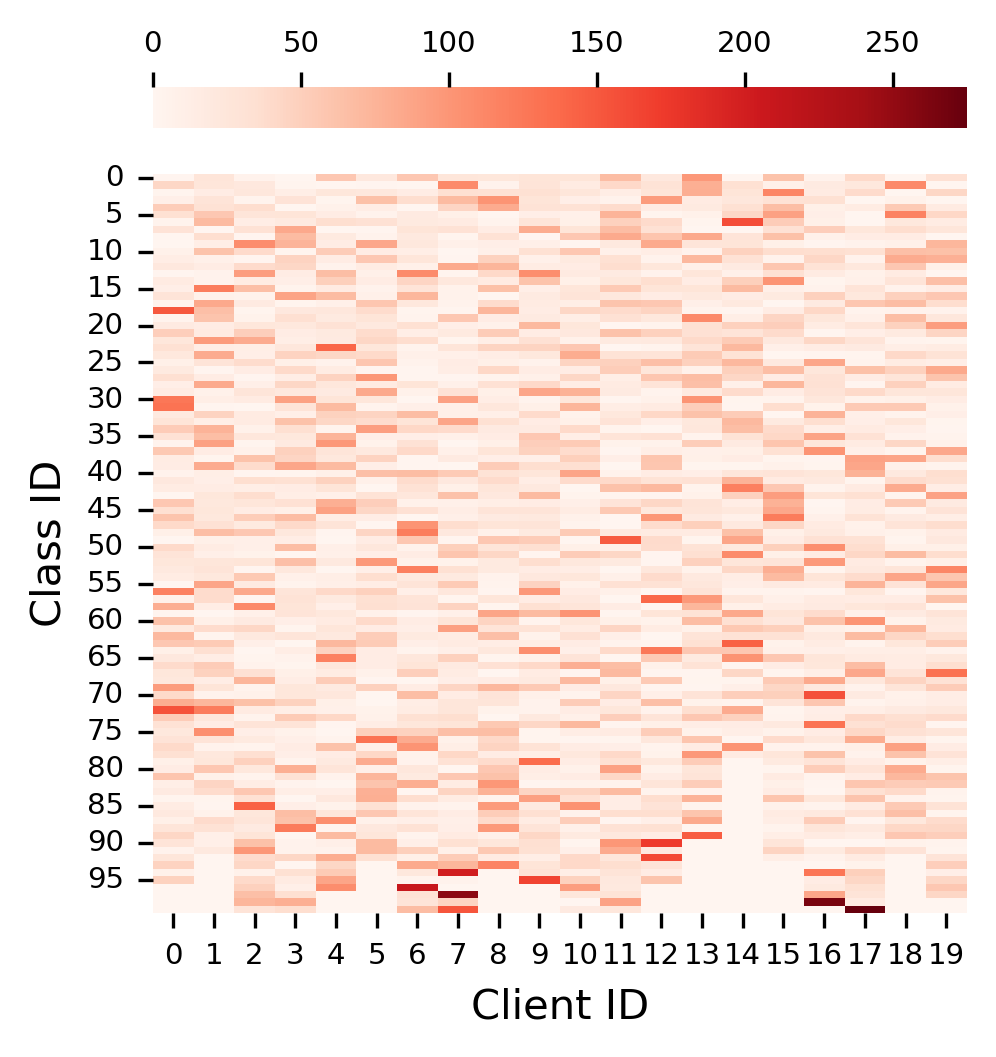}
        \caption{$\alpha=1.0$} 
    \end{subfigure}
    \caption{Data distributions for the CIFAR-100 practical setting.}
    \label{fig:distribution_cifar100}
    \vspace{-10pt}
\end{figure}

Figure \ref{fig:distribution_cifar10} and \ref{fig:distribution_cifar100} show data distributions that vary according to different $\alpha$ values in the Dirichlet distribution, which controls the level of non-IID data partitioning across clients. Each cell in the heatmaps indicates the number of samples per class from each client, with colors representing sample density. 

\section*{D. Communication Efficiency}
The communication costs of FedNH and FedUV scale proportionally with model size since they transmit complete model parameters. 
For FML and FedKD, the communication overhead includes auxiliary network parameters $|\bm{\theta}_{aux}|$ and $|\bm{\phi}_{aux}|$, with FedKD employing dimensionality reduction factor $r$ to mitigate costs. In contrast, PBFL approaches including ProtoNorm achieve significant communication efficiency by transmitting only class prototypes, substantially reducing bandwidth requirements.
\begin{table}[tbh]
    \centering
    {\fontsize{9}{11}\selectfont
    \begin{tabular}{ll}
    \toprule
    Algorithm & Communication cost formulation \\
    \midrule
    FedNH and FedUV & $\sum^M_{i=1} |\bm{w}_i|\times 2$ \\
    LG-FedAvg & $\sum^M_{i=1} |\bm{\phi}_i|\times 2$ \\
    FML & $M\times (|\bm{\theta}_{aux}| + |\bm{\phi}_{aux}|)\times 2$ \\
    FedKD & $M\times (|\bm{\theta}_{aux}| + |\bm{\phi}_{aux}|)\times 2\times r$ \\
    FedDistill & $\sum^M_{i=1} (K_i + K) \times K $ \\
    PBFL approaches & $\sum^M_{i=1} (K_i + K) \times d$ \\
    \bottomrule
    \end{tabular}
    }
    \caption{
    Per-round communication cost formulation.
    }
    \label{table:comm_cost_theory}
\end{table}

\end{document}